\documentclass{article}

\PassOptionsToPackage{numbers, compress}{natbib}

\usepackage[preprint]{neurips_2022}

\usepackage{microtype}
\usepackage{graphicx}
\usepackage{subfigure}
\usepackage{booktabs} 
\usepackage{algorithm}
\usepackage[noend]{algorithmic}
\usepackage{amsmath,amsthm,amsfonts,amssymb}
\usepackage{amsmath}
\usepackage{bbm}
\usepackage{color}
\usepackage{xcolor}
\usepackage{mathrsfs}
\usepackage{bm}
\usepackage{multirow}
\usepackage{booktabs}
\usepackage{makecell}
\usepackage{graphicx}
\usepackage{graphicx}
\usepackage{subfigure}
\usepackage{caption}
\usepackage{thmtools}
\usepackage{thm-restate}
\usepackage{makecell}
\usepackage{pdfpages}
\definecolor{light-gray}{gray}{0.85}
\usepackage{colortbl}
\usepackage{hhline}
\usepackage{xspace}
\usepackage[T1]{fontenc}

\newcommand{\brac}[1]{{\left[ #1 \right]}}

\newcommand{\defeq}{\mathrel{\mathop:}=}

\newcommand{\mat}[1]{\ensuremath{\mathbf{#1}}}

\newcommand{\argmin}{\mathop{\rm argmin}}

\newcommand{\trans}{^{\top}}

\newcommand{\E}{\mathbb{E}}
\newcommand{\D}{\mathbb{D}}
\renewcommand{\P}{\mathbb{P}}

\newcommand{\R}{\mathbb{R}}

\newcommand{\A}{\mat{A}}

\newcommand{\cM}{\mathcal{M}}

\newcommand{\cD}{\mathcal{D}}

\newcommand{\cS}{\mathcal{S}}
\newcommand{\cA}{\mathcal{A}}
\newcommand{\cB}{\mathcal{B}}

\newenvironment{proof-sketch}{\noindent{\bf Proof Sketch}
  \hspace*{1em}}{\qed\bigskip\\}
\newenvironment{proof-idea}{\noindent{\bf Proof Idea}
  \hspace*{1em}}{\qed\bigskip\\}
\newenvironment{proof-of-lemma}[1][{}]{\noindent{\bf Proof of Lemma {#1}}
  \hspace*{1em}}{\qed\bigskip\\}
\newenvironment{proof-of-proposition}[1][{}]{\noindent{\bf
    Proof of Proposition {#1}}
  \hspace*{1em}}{\qed\bigskip\\}
\newenvironment{proof-of-theorem}[1][{}]{\noindent{\bf Proof of Theorem {#1}}
  \hspace*{1em}}{\qed\bigskip\\}
\newenvironment{inner-proof}{\noindent{\bf Proof}\hspace{1em}}{
  $\bigtriangledown$\medskip\\}
\newenvironment{proof-attempt}{\noindent{\bf Proof Attempt}
  \hspace*{1em}}{\qed\bigskip\\}

\usepackage{times}
\usepackage{diagbox}
\usepackage{booktabs, caption, makecell}

\usepackage{threeparttable}
\usepackage{wrapfig}

\newtheorem{theorem}{Theorem}

\theoremstyle{definition}

\newcommand{\alglinelabel}{%
  \addtocounter{ALC@line}{-1}
  \refstepcounter{ALC@line}
  \label
}

\newcommand{\MG}{{\rm MG}}

\newcommand{\reb}[1]{{\color{black}{#1}}}

\newcommand{\nd}{\textsc{Nash\_DQN}\xspace}
\newcommand{\nde}{\textsc{Nash\_DQN\_Exploiter}\xspace}
\newcommand{\nv}{\textsc{Nash\_VI}\xspace}
\newcommand{\nve}{\textsc{Nash\_VI\_Exploiter}\xspace}
\newcommand{\golf}{\textsc{Golf\_with\_Exploiter}\xspace}
\newcommand{\minibatch}{{\mathcal{M}}}
\usepackage{enumitem}
\usepackage[utf8]{inputenc} 
\usepackage[T1]{fontenc}    
\usepackage{hyperref}       
\usepackage{url}            
\usepackage{booktabs}       
\usepackage{amsfonts}       
\usepackage{nicefrac}       
\usepackage{microtype}      
\usepackage{xcolor}         
\usepackage{cleveref}

\title{A Deep Reinforcement Learning Approach for Finding Non-Exploitable Strategies in Two-Player Atari Games}

\author{%
    Zihan Ding\thanks{The first two authors contributed equally.}\\
    Princeton University\\
    \texttt{zihand@princeton.edu}
   \And
  Dijia Su\footnotemark[1] \\
  Princeton University\\
  \texttt{dsu@princeton.edu} \\
  \And
  Qinghua Liu \\
  Princeton University\\
  \texttt{qinghual@princeton.edu} \\
    \And
  Chi Jin \\
  Princeton University\\
  \texttt{chij@princeton.edu} \\
}

\begin{document}

\maketitle

\begin{abstract}
This paper proposes new, end-to-end deep reinforcement learning algorithms for learning two-player zero-sum \emph{Markov games}. 
Different from prior efforts on training agents to beat a fixed set of opponents, our objective is to find the \emph{Nash equilibrium} policies that are free from exploitation by even the adversarial opponents.
We propose (a) \nd algorithm, which integrates the deep learning techniques from single DQN \citep{mnih2013playing} into the classic Nash Q-learning algorithm \citep{hu2003nash} for solving tabular Markov games; (b) \nde algorithm, which additionally adopts an exploiter to guide the exploration of the main agent. 
We conduct experimental evaluation on tabular examples as well as various two-player Atari games. Our empirical results demonstrate that (i) the policies found by many existing methods including NFSP \cite{heinrich2016deep} and PSRO \cite{lanctot2017unified} can be prone to exploitation by adversarial opponents; (ii) the output policies of our algorithms are robust to exploitation, and thus outperform existing methods.
\end{abstract}

\section{Introduction}
\label{intro}

Reinforcement learning (RL) in multi-agent systems has succeeded in many challenging tasks, including Go \citep{go}, hide-and-seek \citep{baker2019emergent}, Starcraft \citep{vinyals2017starcraft},  Dota \citep{berner2019dota}, Poker \citep{heinrich2016deep, brown2019superhuman, zha2021douzero}, and board games \citep{lanctot2019openspiel, serrino2019finding}. 
Except for Poker systems, a large number of these works measure their success in terms of performance against fixed agents, average human players or experts in a few shots. Nevertheless, a distinguishing feature of games is that the opponents can further model the learner's behaviors, adapt their strategies, and exploit the learner's weakness. It is highly unclear whether the policies found by many of these multi-agent systems remain viable against the adversarial exploitation of the opponents.
\begin{figure*}[htbp]
    \centering
    \includegraphics[width=\textwidth]{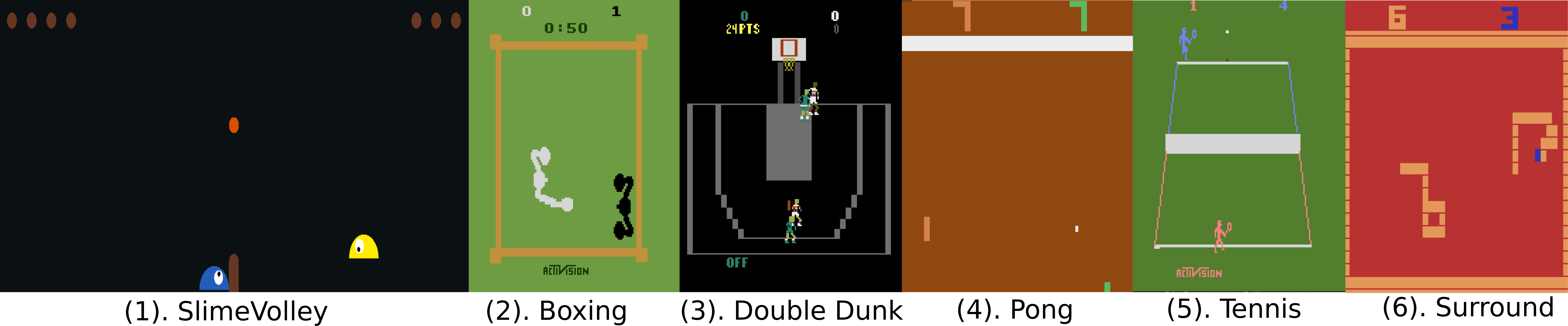}
    \caption{Screen shots of the six two-player video games.}
    \label{fig:games}
\end{figure*}

In this paper, we consider two-player zero-sum \emph{Markov games} (MGs), and our objective is to find the \emph{Nash Equilibrium} (NE)~\citep{nash1950equilibrium}. By definition, the NE strategy is a stationary point where no player has the incentive to deviate from its current strategy. Due to the minimax theorem, the NE strategy for one player is also the best solution when facing against the best response of the opponent. That is, NE in two-player zero-sum games is a natural solution that is free from the exploitation by adversarial opponents.

The concepts of Nash equilibrium and non-exploitability have been well studied in the community of learning \emph{extensive-form games} (EFGs) such as Poker \citep{heinrich2016deep, brown2019superhuman, zha2021douzero, mcaleer2021xdo}.
However, directly applying EFGs algorithms to solve MGs requires us to first convert MGs with general transitions to EFGs with tree-structured transitions. The conversion will lead to an exponential blow-up in the game size in the worst case (especially when there are many states that can be reached by a large number of different paths), resulting in significant inefficiency in practice.
Another line of prior works \citep{heinrich2016deep, lanctot2017unified} directly combine the best-response-based algorithm for finding NE in normal-form games, such as fictitious play~\citep{brown1951iterative} and double oracle~\citep{mcmahan2003planning}, with single-agent deep RL algorithm such as DQN~\citep{mnih2013playing} and PPO~\citep{schulman2017proximal} for finding the best-response. While these approaches are designed for solving MGs, they do not utilize the detailed game-theoretic structure within MGs beyond treating it as normal-form games, which will again lead to the significant inefficiency in learning (see detailed theoretical arguments in Section \ref{sec:related}, and empirical performance on tabular and Atari environments in Section \ref{sec:exp}).

This paper proposes two new, end-to-end deep reinforcement learning algorithms for learning the Nash equilibrium of two-player zero-sum MGs: Nash Deep Q-Network (\nd), and its variant Nash Deep Q-Network with Exploiter (\nde).  
\nd  effectively combines (a) the recent deep learning techniques DQN \citep{mnih2013playing} for addressing single-agent RL with continuous state space and function approximation, with (b) the classic algorithms Nash VI or Nash Q-learning \cite{hu2003nash} for solving tabular Markov games.
\nde is a variant of \nd which explicitly train an exploiting opponent during its learning. The exploiting opponent stimulates the exploration for the main agent. Both algorithms are the practical variants of theoretical algorithms which are guaranteed to efficiently find Nash equilibria in the basic tabular setting.


Experimental evaluations are conducted on both tabular MGs and two-player video games, to show the effectiveness and robustness of the proposed algorithms. As shown in Fig.~\ref{fig:games}, the video games in our experiments include five two-player Atari games in PettingZoo library \citep{pettingzoo, Bellemare_2013} and a benchmark environment Slime Volley-Ball \citep{slimevolleygym}. Due to the constraints of computational resource, we consider the RAM-based version of Atari games, and truncate the length of each game to 300 steps. We test the performance by training adversarial opponents using DQN that directly exploit the learner's policies. Our experiments in both settings show that our algorithms significantly outperform standard algorithms for MARL including Neural Fictitious Self-Play (NFSP) \citep{heinrich2016deep} and Policy Space Response Oracle (PSRO) \citep{lanctot2017unified}, in terms of the robustness against adversarial exploitation.
The code for experiments is released.\footnote{The repositories for code implementation are released: 1. \href{https://github.com/quantumiracle/nash-dqn}{https://github.com/quantumiracle/nash-dqn} is an independent implementation of our proposed algorithms; 2. \textit{MARS} (\href{https://github.com/quantumiracle/MARS}{https://github.com/quantumiracle/MARS}) is a library for MARL algorithm evaluation.}

Even though our \nd algorithm is a natural integration of existing deep learning techniques and classic algorithms for MGs, to our best knowledge, this natural algorithm has never been formally proposed before. Neither has it been thoroughly tested, and compared against existing popular algorithms in representative environments such as Atari games, especially in terms of robustness to exploitation. We believe there lies the main contribution of this paper, to fill this important missing piece in the literature.

\section{Related Works}
\label{sec:related}
\textbf{MARL in IIEFGs.} This line of work \cite{lanctot2009monte,  heinrich2015fictitious, brown2019superhuman, farina2020stochastic, mcaleer2021xdo, kozuno2021model, bai2022near} focuses on learning Imperfect Information Extensive-Form Games (IIEFGs). This line of results focuses on finding Nash equilibrium, which is non-exploitable.
However, most algorithms designed for IIEFGs heavily rely on the perfect recall property---each information set can be only reached by one path of history. This is rather different from MGs, where states can be reached potentially a large number of different paths (as in the case of many Atari games). As a result, directly applying the IIEFG solutions to MGs requires us to first convert MGs into IIEFGs, which can lead to an exponential blow-up in the game size as well as significant inefficiency in practice.

\textbf{MARL in zero-sum MGs (empirical results).}
In addition to IIEFG algorithms, there is another line of empirical algorithms that can be applied to solve zero-sum MGs.
These algorithms involve combining single-agent RL algorithm, such as deep Q-network (DQN)~\citep{mnih2013playing} or proximal policy optimisation (PPO)~\citep{schulman2017proximal}, with best-response-based Nash equilibrium finding algorithm for \emph{Normal-Form} Games (NFGs), including fictitious play (FP)~\citep{brown1951iterative}, double oracle (DO)~\citep{mcmahan2003planning}, and many others. A few other examples such as neural fictitious self-play (NFSP)~\citep{heinrich2016deep}, policy space response oracles (PSRO)~\citep{lanctot2017unified}, online double oracle~\citep{dinh2021online} and prioritized fictitious self-play \citep{vinyals2019grandmaster} also fall into this general class of algorithms or their variants.
These algorithms call single-agent RL algorithm to compute the best response of the current ``meta-strategy'', and then use the best-response-based NFG algorithm to compute a new ``meta-strategy''. However, these algorithms inherently treat MGs as NFGs, do not efficiently utilize the fine game-theoretic structure within MGs. In particular, for NFGs, FP has a convergence rate exponential in the number of actions, while DO has a linear rate. A direct conversion of Markov game to normal-form game will generate a new NFG whose action size scales
\emph{exponentially} in horizon, number of states and actions in the original MGs. This, as shown in our experiments, leads to significant inefficiency in scaling up with size of the MGs.

We remark that a recent paper \cite{bakhtin2021no} proposes Deep Nash Value Iteration algorithm for solving no-press diplomacy game. Despite of a similar name and a similar starting point based on Nash Q-learning, their algorithm is rather different from our \nd in the following critical perspectives: while previous work~\cite{bakhtin2021no} additionally handles the challenge of large action space, the key update of their algorithm further requires (a) deterministic transition, (b) reward only depends on the next state, and most importantly (c) transition is \emph{known}. These restrictions prevent their algorithms from applying to general MGs which are the focus of this paper. In particular, learning from environments with \emph{unknown}  transition function is one of the most important challenge in addressing Atari games. 

\textbf{MARL in zero-sum MGs (theory).}
There has been rich studies on two-player zero-sum MGs from the theoretical perspectives. Many of these works \cite{hu2003nash, bai2020provable, bai2020near, liu2021sharp, jin2021v} focused on the tabular setting, which requires the numbers of states and actions to be finite. These algorithms are proved to converge to the NE policies in a number of samples that is \textbf{polynomial} in the number of states, actions, horizon (or the discount coefficient), and the target accuracy. Among those, Nash Q-learning \citep{hu2003nash} is one of the earliest works along this line of research, which provably converges to NE for general-sum games under the assumption that the NE is unique for each stage game during the learning process. On the other hand, \golf \citep{jin2021power} is another theoretical work with provable polynomial convergence for two-player zero-sum MGs. Our \nd algorithm is designed based on the provable tabular algorithm Nash Value Iteration \citep{liu2021sharp}, which is a natural extension of value iteration algorithm from single-agent setting to the multi-agent setting. For better understanding, we provide a detailed comparison of similarities and differences of Nash Q-learning, Nash Value Iteration, \golf and \nd in the Appendix \ref{app:sec_compare_tabular_algs}. \cite{xie2020learning} considers MGs with linear function approximation.  There are a few theoretical works on studying zero-sum MGs with general function approximation \citep{jin2021power, huang2021towards}, which include neural network function approximation as special cases. However, these algorithms are sample-efficient, but not computationally efficient. They require solving optimistic policies with complicated confidence sets as constraints, which cannot be run in practice.

\section{Preliminaries} \label{sec:prelim}
In this paper, we consider Markov Games \citep[MGs,][]{shapley1953stochastic, littman1994markov}, which generalizes standard Markov Decision Processes (MDPs) into  multi-player settings. Each player has its own utility and optimize its policy to maximize the utility. We consider a special setting in MG called two-player zero-sum games, which has a competitive relationship between the two players. 


More concretely, consider a infinite-horizon discounted version of two-player zero-sum MG, which is denoted as $\MG(\cS, \cA, \cB, \P, r, \gamma)$. $\mathcal{S}$ is the state space, $\mathcal{A}$ and $\mathcal{B} $ are the action spaces for the max-player and min-player respectively. $\P$ is the state transition distribution, and $\P ( \cdot | s, a, b) $ is the distribution of the next state given the current state $s$ and action pair $(a, b)$. $r\colon \cS \times \cA \times \cB \to \mathbb{R}$ is the reward function. In the zero-sum setting, the reward is the gain for the max-player and the loss for the min-player due to the zero-sum payoff structure. $\gamma\in[0,1]$ is the discount factor. 
At each step, the two players will observe the state $s \in \cS$ and choose their actions $a \in \cA$ and $b \in \cB$ independently and simultaneously. The action from their opponent can be observed after they take the actions, and the reward $r(s, a, b)$ will be received ($r$ for max-player and $-r$ for min-player). The environment then transit to the next state $s'\sim\P(\cdot | s, a, b)$.



\paragraph{Policy, value function.}
We define the policy and value functions for each player. For the max-player, the (Markov) policy is a map $\mu: \cS \rightarrow \Delta_{\cA} $. Here we only consider discrete action space, so $\Delta_{\cA}$ is the probability simplex over action set $\cA$. Similarly, the policy for the min-player is $\nu: \cS \rightarrow \Delta_{\cB} $. 


$V^{\mu, \nu} \colon \cS \to \mathbb{R}$ represents the value
function evaluated with policies $\mu$ and $\nu$, which can be expanded as the expected cumulative reward starting from the state $s$, 
\begin{equation} \label{eq:V_value}
\begin{aligned}
	 V^{\mu, \nu}(s) 
\defeq \E_{\mu, \nu}\bigg[\sum_{h =
        1}^\infty \gamma^{h-1} r(s_{h}, a_{h}, b_{h}) \bigg| s_1 = s\bigg].
\end{aligned}
\end{equation}
Correspondingly, $Q^{\mu, \nu}:\cS \times \cA \times \cB \to \mathbb{R}$ is the state-action value function evaluated with policies $\mu$ and $\nu$, which can also be expanded as expected cumulative rewards as:
\begin{equation} \label{eq:Q_value}
\begin{aligned}
  & \quad Q^{\mu, \nu}(s, a, b) \defeq \E_{\mu,
    \nu}\bigg[\sum_{h = 1}^\infty \gamma^{h-1}r(s_{h},  a_{h}, b_{h})
  \big| s_1 = s, a_1 = a, b_1 = b\bigg].
\end{aligned}
\end{equation}
In this paper we also use a simplified notation for convenience, $[\P V](s, a, b) \defeq \E_{s' \sim \P(\cdot|s, a,
  b)}V(s')$, where $\P$ as the transition function can be viewed as an operator. Similarly, we denote  $[\D_\pi Q](s) \defeq \E_{(a, b) \sim \pi(\cdot, \cdot|s)} Q(s, a, b)$ for any state-action value function. In this way, the Bellman equation for two-player MG can be written as:
\begin{equation}
 Q^{\mu, \nu}(s, a, b) =
  (r + \gamma \P V^{\mu, \nu})(s, a, b), \quad    V^{\mu, \nu}(s)
  =  (\D_{\mu\times\nu} Q^{\mu, \nu})(s),
\label{eq:bellman}
\end{equation}
for all $(s, a, b) \in \cS \times \cA \times \cB$.



\paragraph{Best response and Nash equilibrium.}
In two-player games, if the other player always play a fixed Markov policy, optimizing over learner's policy is the same as optimizing over the policy of single agent in MDP (with other players' polices as a part of the environment). For two-player cases, given the max-player's policy $\mu$, there exists a \emph{best response} of the min-player, which is a policy
$\nu^\dagger(\mu)$ satisfying $V^{\mu, \nu^\dagger(\mu)}(s) = \inf_{\nu} V^{\mu, \nu}(s)$ for
any $s \in \cS$. We simplify the notation as: $V^{\mu, \dagger} \defeq V^{\mu, \nu^\dagger(\mu)}$. Similar best response for a given min-player's policy $\nu$ also exists as $\mu^\dagger(\nu)$ satisfying $V^{\dagger, \nu}=\sup_\mu V^{\mu, \nu}$. By leveraging the Bellman equation Eq.~\eqref{eq:bellman}, the best response can be derived with dynamic programming,
\begin{equation}
\label{eq:best_response_v}
    Q^{\mu, \dagger}(s,a,b) = (r+\gamma \mathbb{P}V^{\mu, \dagger})(s,a,b), \quad 
    V^{\mu, \dagger}(s) = \inf_{\nu}(\mathbb{D}_{\mu\times \nu}Q^{\mu, \dagger})(s)
\end{equation}

The \emph{Nash equilibrium} (NE) is defined as a pair of policies $(\mu^\star,\nu^\star)$ serving as the optimal against the best responses of the opponents, indicating:
\begin{equation}
\begin{aligned}
	  V^{\mu^\star, \dagger}(s) = \sup_{\mu}
      V^{\mu, \dagger}(s), \quad
       V^{\dagger, \nu^\star}(s) = \inf_{\nu}
      V^{\dagger, \nu}(s),\end{aligned}
\end{equation}
for all $s \in \cS$. The existence of NE is shown by previous work~\cite{filar2012competitive}. Furthermore, NE strategies satisfy the following minimax equation:
\begin{equation}
\textstyle \sup_{\mu} \inf_{\nu} V^{\mu, \nu}(s) = V^{\mu^\star, \nu^\star}(s) = \inf_{\nu} \sup_{\mu} V^{\mu, \nu}(s).
\end{equation}
which is similar as the normal-form game but without the bilinear structure of the payoff matrix. NE strategies are the ones where no player has incentive to change its own strategy. The value functions of $(\mu^\star,\nu^\star)$ is denoted as $V^{\star}$ and $Q^\star$, which satisfy the following Bellman optimality equation:
\begin{equation}
\label{eq:nash_q}
\begin{aligned}
Q^{\star}(s,a,b) &= (r+\gamma\mathbb{P}V^{\mu, \dagger})(s,a,b)  \\
V^{\star}(s) =
  \sup_{\mu \in \Delta_{\cA}}\inf_{\nu \in \Delta_{\cB}} &(\D_{\mu \times \nu} Q^{\star})(s)
  = \inf_{\nu \in \Delta_{\cB}}\sup_{\mu \in \Delta_{\cA}} (\D_{\mu \times \nu} Q^{\star})(s).
\end{aligned}
\end{equation}

\paragraph{Learning Objective.}
The \emph{exploitability} of policy $(\hat{\mu}, \hat{\nu})$ can be defined as the difference in values comparing to Nash strategies when playing against their best response. Formally, the exploitability of the max-player can be defined as $V^{\star}(s_1) -   V^{\hat{\mu},\dagger}(s_1)$ while the exploitability of the min-player is defined as $V^{\dagger, \hat{\nu}}(s_1) - V^{\star}(s_1)$. We define the total suboptimality of $(\hat{\mu}, \hat{\nu})$ simply as the summation of the exploitability of both players
\begin{equation}
\label{eq:subopt}
  V^{\dagger, \hat{\nu}}(s_1) - V^{\hat{\mu}, \dagger}(s_1) 
	  = \brac{V^{\dagger, \hat{\nu}}(s_1) - V^{\star}(s_1)} 
	  +  \brac{V^{\star}(s_1) -   V^{\hat{\mu},\dagger}(s_1)}.
\end{equation}
This quantity is also known as the duality gap in the literature of MGs, which can be viewed as a distance measure to Nash equilibria. We note that the duality gap of Nash equilibria is equal to zero. Furthermore, all video games we conduct experiments on are symmetric to two players, which implies that $V^\star(s_1) = 0$.

\paragraph{\nv and Nash Q-Learning.}

The model-based tabular algorithm for MGs---\nv, computes the near-optimal policy by performing Bellman optimality update Eq.~\eqref{eq:nash_q} with $(\P, r)$ replaced by their empirical estimates using samples. The empirical estimates of the entry $\hat{P}(s'|s, a, b)$ in transition matrix is computed by how many times $(s, a, b, s')$ is visited divided by how many times $(s, a, b)$ is visited. Please see Algorithm \ref{alg:sec_nvi} in Appendix for more details.
Similar to the relation between Q-learning and VI in the single-agent setting, 
Nash Q-Learning \citep{hu2003nash} is the model-free version of \nv, which performs incremental update
$$Q(s, a, b)\leftarrow (1-\alpha) Q(s, a, b) + \alpha(r + \gamma \cdot {\rm Nash}(Q(s', \cdot, \cdot))$$
whenever a new sample $(s, a, b, r, s')$ is observed. Please see pseudo-codes and detailed discussions about \nv and Nash Q-Learning in Appendix~\ref{app:sec_nash_q_l}, \ref{app:sec_nvi} and \ref{app:sec_compare_tabular_algs}.

\section{Methodology}

To learn the Nash equilibria of two-player zero-sum Markov games, this paper proposes two novel, end-to-end deep MARL algorithms---\nd and \nde. \nd combines single-agent DQN \citep{mnih2013playing} with \nv \citep{liu2021sharp}---a provable algorithm for tabular Markov games.
\nde is a variant of \nd by explicitly training an adversarial opponent during the learning phase to encourage the exploration of the learning agent. \reb{The details of \nv is discussed in Appendix \ref{app:sec_nvi}.}

\subsection{\nd}
\label{subsec:nash_dqn}

\begin{algorithm}[t]
\caption{Nash Deep Q-Network (\nd)}
\begin{algorithmic}[1]
\STATE Initialize replay buffer $\cD=\emptyset$, counter $i=0$, Q-network $Q_\phi$
\STATE Initialize target network parameters: $\phi^\text{target}\leftarrow \phi$.
\FOR{episode $k=1,\ldots,K$}
\STATE reset the environment and observe $s_1$.
\FOR{$t=1,\ldots,H$}
\STATE {\color{blue}\% collect data}
\STATE sample actions $(a_t, b_t)$ from 
$\begin{cases} 
\text{Uniform}(\cA \times \cB) & \quad \text{with probability~} \epsilon \\
(\mu_t, \nu_t)=\textsc{Nash}(Q_{\phi}(s_t,\cdot, \cdot)) & \quad \text{otherwise.}
\end{cases}$
\STATE execute actions $(a_t, b_t)$, observe reward $r_t$, next state $s_{t+1}$.
\STATE store data sample $(s_t,a_t,b_t,r_t,s_{t+1})$ into $\cD$ 
\STATE {\color{blue}\% update Q-network}
\STATE randomly sample minibatch $\cM \subset \{1, \ldots, |\mathcal{D}|\}$.
\FOR{all $j \in \minibatch$}
\STATE compute $(\hat{\mu}, \hat{\nu})=\textsc{Nash}(Q_{\phi^{\text{target}}}(s_{j+1}, \cdot, \cdot))$
\STATE \alglinelabel{line:target_nash_DQN} set $y_j =  r_j + \gamma \hat{\mu}\trans Q_{\phi^{\text{target}}}(s_{j+1}, \cdot, \cdot) \hat{\nu}$. 
\ENDFOR
 \STATE Perform $m$ steps of GD on loss $\sum_{j \in \minibatch}(y_j-Q_\phi(s_j,a_j,b_j))^2$ to update $\phi$.  \alglinelabel{line:square_loss_nd}
\STATE {\color{blue}\% update target network}
\STATE $i=i+1$; if $i\%N=0$: $\phi^\text{target}\leftarrow \phi$.
\ENDFOR
\ENDFOR	
\end{algorithmic}
\label{alg:nash_dqn}
\end{algorithm}

We describe \nd~in Algorithm \ref{alg:nash_dqn} which incorporates neural networks into the tabular \nv \citep{liu2021sharp} algorithm for approximating the Q-value function. 
Similar to the single-agent DQN, \nd~maintains two networks in the training process: the $Q$-network and its target network, which are parameterized by $\phi$ and $\phi^\text{target}$, respectively.
In each episode, \nd~executes the following two main steps:

$\bullet$ \textbf{Data collection:} \nd adopts the $\epsilon$-greedy strategy for exploration. At each state $s_t$, with probability  $\epsilon$, both players take random actions; otherwise, they will sample actions $(a_t,b_t)$ from the Nash equilibrium of its Q-value matrix (i.e., $\textsc{Nash}(Q_{\phi}(s_t,\cdot, \cdot))$, see Eq.~\eqref{eq:nash_subroutine}). After that, we add the collected data into the experience replay buffer $\cD$.

$\bullet$  \textbf{Model update}: We first randomly sample a batch of data $\cM$ from replay buffer $\cD$, and then perform $m$-steps gradient descent to update $\phi$ using the loss below
$$\textstyle
\sum_{j \in\cM}\left(Q_\phi(s_j,a_j,b_j)-y_j\right)^2,
$$
where target $y_j$ is computed according to line 14. We adopt the convention of setting $Q_{\phi^{\text{target}}}(s_{j+1}, \cdot, \cdot) = 0$ for all terminal state $s_{j+1}$.

Here, $\textsc{Nash}(\cdot)$ is the NE subroutine for normal-form games, which takes a payoff matrix $\A \in \R^{A \times B}$ as input and outputs  one of its Nash equilibria $(\mu^\star, \nu^\star)$. In math, we have:
\begin{equation} \label{eq:nash_subroutine}
    (\mu^\star, \nu^\star) = \textsc{Nash}(\A) \quad\text{if and only if}\quad  \forall \mu, \nu, ~~ \mu\trans \A \nu^\star \le (\mu^\star)\trans \A \nu^\star \le (\mu^\star)\trans \A \nu.
\end{equation}
There are several off-the-shelf libraries to implement this  $\textsc{Nash}$ subroutine.
After comparing the performance of 
several different implementations (see Appendix~\ref{sec:choose_solver} for details),  we found the ECOS library \citep{domahidi2013ecos} works the best, which from now on is set as the default choice in our algorithms. 

Regarding the choice of target value update (line \ref{line:target_nash_DQN}), one can view it as a Monte Carlo estimate of 
\begin{equation}\label{eq:nash_dqn_target_expectation}
   \textstyle
r(s_j,a_j,b_j)+ \gamma\E_{s'\sim\P(\cdot\mid s_j,a_j,b_j)}[ \max_{\hat\mu\in\Delta_A} \min_{\hat\nu\in\Delta_B} \hat\mu^\intercal Q_{\phi^{\text{target}}}(s', \cdot, \cdot)\hat\nu]. 
\end{equation}
Intuitively, we aim to approximate the Q-value function of Nash equilibria $Q^\star$ by our Q-network $Q_\phi$.
Recall that $Q^\star$ is the \emph{unique} solution of the Bellman optimality equations: 
\begin{equation}
   \textstyle \label{eq:Qstar}
\forall (s,a,b), \quad Q^\star(s,a,b) = r(s,a,b) + \gamma\E_{{s}'\sim\P(\cdot\mid s,a,b)}[ \max_{\hat\mu\in\Delta_A} \min_{\hat\nu\in\Delta_B} \hat\mu^\intercal Q^\star({s}', \cdot, \cdot)\hat\nu].
\end{equation}
As a result, by performing gradient descent on $\phi$ to minimize the square loss as in line \ref{line:square_loss_nd}, $Q_\phi$ will decrease its Bellman error, and eventually converge to $Q^\star$, as more samples are collected. Finally, we remark that the target Nash $Q$-network ($Q_{\phi^{\text{target}}}$) is updated in a delayed manner as DQN to stabilize the training process.

\subsection{\nde}
\label{subsec:nash_dqn_exploiter}

\begin{algorithm}[t]
\caption{Nash Deep Q-Network with Exploiter (\nde)}
\begin{algorithmic}[1]
\STATE Initialize replay buffer $\cD=\emptyset$, counter $i=0$, Q-network $Q_\phi$, exploiter network $\tilde{Q}_{\psi}$.
\STATE Initialize target network parameters: $\phi^\text{target}\leftarrow \phi$, $\psi^\text{target}\leftarrow \psi$.
\FOR{episode $k=1,\ldots,K$}
\STATE reset the environment and observe $s_1$.
\FOR{$t=1,\ldots,H$}
\STATE {\color{blue}\% collect data}
\STATE sample actions $(a_t, b_t)$ from 
$\begin{cases} 
\text{Uniform}(\cA \times \cB) & \quad \text{with probability~} \epsilon \\
(\mu_t, \nu_t) \text{~computed according to Eq.~\eqref{eq:compute_policy}} & \quad \text{otherwise.}
\end{cases}$
\STATE execute actions $(a_t, b_t)$, observe reward $r_t$, next state $s_{t+1}$.
\STATE store data sample $(s_t,a_t,b_t,r_t,s_{t+1})$ into $\cD$.
\STATE {\color{blue}\% update Q-network and exploiter network}
\STATE randomly sample minibatch $\minibatch \subset \{1, \ldots, |\cD|\}$.
\FOR{all $j \in \mathcal{M}$}
\STATE compute $(\hat{\mu}, \hat{\nu})=\textsc{Nash}(Q_{\phi^{\text{target}}}(s_{j+1}, \cdot, \cdot))$
\STATE set $y_j =  r_j + \gamma \hat{\mu}\trans Q_{\phi^{\text{target}}}(s_{j+1}, \cdot, \cdot)\hat{\nu}$.
\STATE set $\tilde{y}_j = r_j+ \gamma \min_{b\in\mathcal{B}}\hat{\mu}\trans \tilde Q_{\psi^\text{target}}(s_{j+1},\cdot,b)$
\alglinelabel{line:exploiter_target}
\ENDFOR
\STATE Perform $m_1$ steps of GD on loss $\sum_{j \in \minibatch}(y_j-Q_\phi(s_j,a_j,b_j))^2$ to update $\phi$.
\STATE Perform $m_2$ steps of GD on loss $\sum_{j \in \minibatch}(\tilde{y}_j-\tilde Q_\psi(s_j,a_j,b_j))^2$ to update $\psi$. \alglinelabel{line:square_loss_nde}
\STATE {\color{blue}\% update target network}
\STATE $i=i+1$; if $i\%N=0$: $\phi^\text{target}\leftarrow \phi$, $\psi^\text{target}\leftarrow \psi$.
\ENDFOR
\ENDFOR	
\end{algorithmic}
\label{alg:nash_dqn_exp}
\end{algorithm}

\nd relies on  the $\epsilon$-greedy strategy for exploration. To improve the exploration efficiency, we  propose a variant of \nd --- \nde, which additionally introduces an exploiter in the training procedure. 
By constantly exploiting the weakness of the main agent, the exploiter forces the main agent to play the part of the games she is still not good at, and thus helps the main agent improve and discover more effective strategies.

We describe \nde in Algorithm \ref{alg:nash_dqn_exp}. We let the main agent maintain a Q-network $Q_{\phi}$ and let the exploiter maintain a separate value network $\tilde{Q}_{\psi}$, both are functions of $(s, a, b)$.
We make two key modifications from \nd. First, in the data collection phase, at state $s_t$, we no longer choose both $\mu_t,\nu_t$ to be the Nash equilibrium computed from $Q_\phi$.
Instead, we only choose $\mu_t$ to be the Nash strategy of $Q_\phi$ but pick the policy of the exploiter $\nu_t$ to be the best response of $\mu_t$ under the exploiter's $Q$-network $\tilde{Q}_{\psi}$. Formally,
\begin{align}\label{eq:compute_policy}
&(\mu_t, \cdot) =  \textsc{Nash}(Q_{\phi}(s_t,\cdot, \cdot))\\
&\nu_t = \argmin_\nu \mu_t\trans\tilde{Q}_{\psi}(s_t, \cdot, \cdot) \nu. \nonumber
\end{align}
In the model update phase, \nde follows exactly the same rule as \nd to update $Q_\phi$ and $Q_{\phi^\text{target}}$, the $Q$-networks of the main agent.
However, for the update of the exploiter networks, \nde utilizes a different regression target in the loss function as specified in line \ref{line:exploiter_target}.
One can view the target as a Monte Carlo estimate of 
 \begin{equation}
    \begin{aligned}\textstyle
    & r(s_j,a_j,b_j)+\gamma \E_{s'\sim\P(\cdot\mid s_j,a_j,b_j)}\left[ \min_b \hat\mu(s')^\intercal \tilde{Q}_{\psi^{\text{target}}}(s', \cdot, b)\right]
        \label{eq:nash_dqn_exploiter_target_expectation}
        \end{aligned}
\end{equation}
We set the target in this way because we aim to approximate $Q^{\hat\mu,\dagger}$, which is the value of the current policy of the main player $\hat{\mu}$ against its best response, using our exploiter network $\tilde{Q}_{\psi}$.
Recall that $Q^{\hat\mu,\dagger}$  satisfies the following  Bellman equations for the best response:
\begin{equation}
   \textstyle \label{eq:Qdagger}
   \forall (s,a,b), \quad Q^{\hat\mu,\dagger}(s,a,b) = r(s,a,b) + \gamma\E_{{s}'\sim\P(\cdot\mid s,a,b)}[  \min_{b} \hat\mu(s')^\intercal Q^{\hat\mu,\dagger}({s}', \cdot, b)].
\end{equation}
Therefore, by performing gradient descent on $\psi$ to minimize the square loss as in line \ref{line:square_loss_nde}, $\tilde{Q}_\psi$ will decrease its (best response version of) Bellman error, and approximate $Q^{\hat\mu,\dagger}$.

\subsection{Theoretical Justification}
\label{sec:theo_just}

With special choices of Q-network architecture $Q_\phi$, minibatch size $|\minibatch|$ and number of steps for GD $m$, both our algorithms \nd and \nde reduce to the $\epsilon$-greedy version of standard algorithm \nv \citep{liu2021sharp} and \nve \citep{jin2021power} for learning tabular Markov games\footnote{With an alternative choice of $(|\minibatch|, m)$, we can also reduce \nd to Nash Q-learning in the tabular setting. For simplicity, we stick to the \nv viewpoint here. Result similar to Theorem \ref{thm:theory} also holds for Nash Q-learning \cite{bai2020near}.}, where the numbers of states and actions are finite and small. Please see Appendix \ref{sec:alg_tab_markov} for a detailed discussion on the connections of these algorithms.

When replacing the $\epsilon$-greedy exploration with optimistic exploration (typically in the form of constructing upper confidence bounds), both \nv and \nve are guaranteed to efficiently find the Nash equilibria of MGs in the tabular settings.

\begin{theorem}[\citep{liu2021sharp,jin2021power}] 
\label{thm:theory}For tabular Markov games, the optimistic versions of both \nv and \nve can find $\epsilon$-approximate Nash equilibria in $\text{poly}(S, A, B, (1-\gamma)^{-1}, \epsilon^{-1}, \log(1/\delta))$ steps with probability at least $1-\delta$. Here $S$ is the size of states, $A, B$ are the size of two players' actions respectively, and $\gamma$ is the discount factor.
\end{theorem}

We defer the proof of Theorem \ref{thm:theory} to Appendix \ref{sec:alg_tab_markov}. We highlight that, in contrast, existing deep MARL algorithms such as NFSP \citep{heinrich2016deep} or PSRO \citep{lanctot2017unified} are incapable of efficient learning of Nash equilibria with a polynomial convergence rate for tabular Markov games.
Our simulation results reveal that they are indeed highly inefficient in finding Nash equilibria.

\section{Experiments}  \label{sec:exp}
The experimental evaluations are conducted on randomly generated tabular Markov games and two-player video games on Slime Volley-Ball \citep{slimevolleygym} and PettingZoo Atari~\citep{pettingzoo}. We tested the performance of proposed methods as well as the baseline algorithms in both (a) the basic tabular form without function approximation (only in tabular environments); and (b) full versions with deep neural networks (in both tabular environments and video games). For (a), we measure the exploitability by computing the exact best response using the ground truth transition and reward function. This is only feasible in the tabular environment. For (b), we measure the exploitability by training single-agent DQN (exploiter) against the learned policy to directly exploit it.

\subsection{Baselines}
For benchmarking purpose, we have the following baselines with deep neural network function approximation for scalable tests:
\begin{itemize}[noitemsep,topsep=0pt, leftmargin=*]
    \item \textbf{Self-Play} (SP): each agent learns to play the best response strategy against the fixed opponent strategy alternatively, i.e., iterative best response.
    \item \textbf{Fictitious Self-Play} (FSP)~\cite{heinrich2015fictitious}: each agent learns a best-response strategy against the episodic average of its opponent's historical strategy set, and save it to its own strategy set. 
    \item \textbf{Neural Fictitious Self-Play} (NSFP)~\cite{heinrich2016deep}: an neural network approximation of FSP, a policy network is explicitly maintained to imitate the historical behaviours by an agent, and the learner learns the best response against it.  
    \item \textbf{Policy Space Response Oracles} (PSRO)~\cite{lanctot2017unified}: we adopt a version based on double oracle (DO), each agent learns the best response against a meta-Nash strategy of its opponent's strategy set, and add the learned strategy to its own strategy set.
\end{itemize}

For tabular case without function approximation, SP, FSP and DO are implemented with Q-learning as the base learning agents for finding best responses.
For tabular case with function approximation and video games, all four baseline methods use DQN as the basic agent for learning the best-response strategies. The pseudo-codes for algorithms SP, FSP, DO are provided in Appendix~\ref{sec:alg_tab_markov}.

\subsection{Tabular Markov Game}
\textbf{Tabular forms without function approximation.}
We first evaluate methods (1) SP, (2) FSP, (3) PSRO, (4) \nd and (5) \nde without function approximation (i.e., w/o neural network) on the tabular Markov games. They reduce to methods (1) SP, (2) FSP, (3) DO, (4) Nash value iteration (\nv) and (5) Nash value iteration with exploiter (\nve), correspondingly.  For SP, FSP and DO, we adopt Q-learning as a subroutine for finding the best response policies. As tabular versions of our deep MARL algorithms, \nv and \nve also use $\textsc{Nash}$ subroutine for calculating NE in normal-form games, with $\epsilon=0.5$ for $\epsilon$-greedy exploration. The pseudo-codes for \nv and \nve are provided in
Appendix \ref{sec:alg_tab_markov}.

We randomly generated the tabular Markov games, which has discrete state space $\mathcal{S}$, discrete action spaces $\mathcal{A}, \mathcal{B}$ for two players and the horizon $H$\footnote{We encode the horizon into the state space in order to use the algorithm designed for the discounted setting.}
The state transition probability function $\{\mathcal{T}_h:\mathcal{S}\times \mathcal{A} \times \mathcal{S}\rightarrow [0,1], h\in[H]\}$ and reward function $\{R_h:\mathcal{S}\times \mathcal{A} \times \mathcal{S}\rightarrow [-1,1], h\in[H]\}$ are both i.i.d sampled uniformly over their ranges.
\begin{figure}[htbp]
    \centering
    \includegraphics[width=0.45\columnwidth]{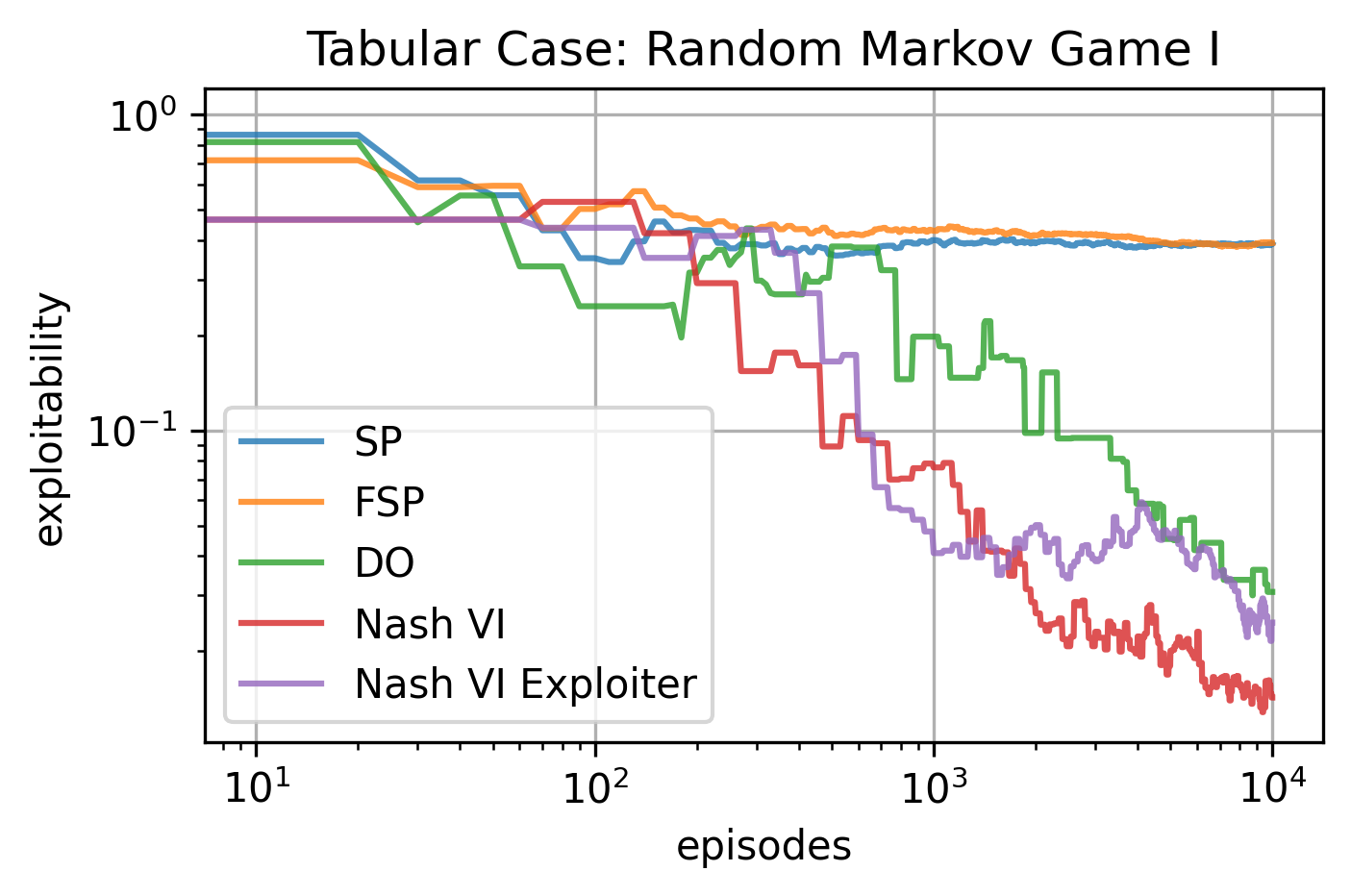}
    \hspace{0.1in}
    \includegraphics[width=0.45\columnwidth]{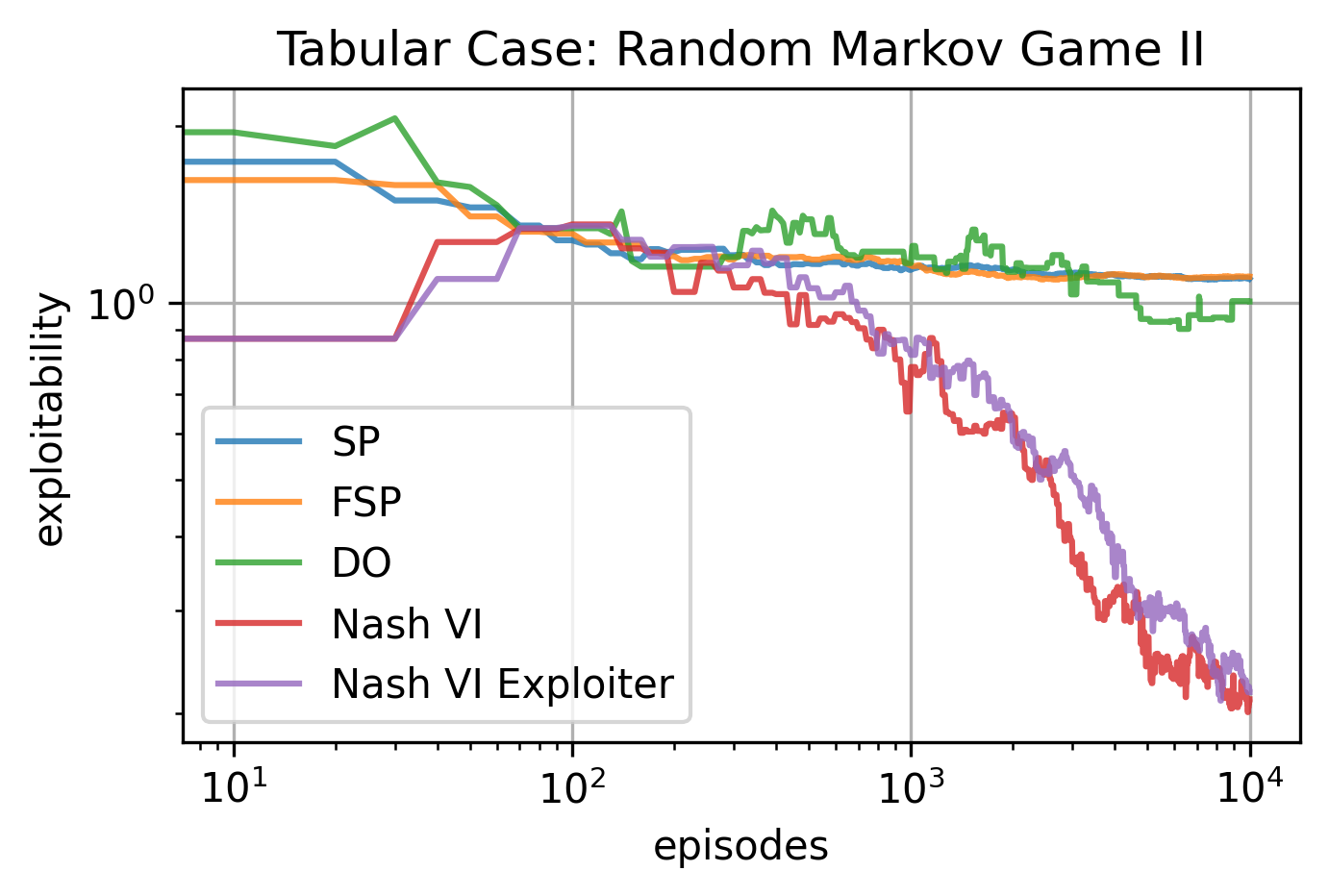}
    \caption{Tabular case experiments on two randomly generated Markov games.}
    \label{fig:tabular_nvi}
    \vskip -.05in
\end{figure}

As shown in Fig.~\ref{fig:tabular_nvi}, we tested on two randomly generated Markov games of different sizes: I. $|\mathcal{S}|=|\mathcal{A}|=|\mathcal{B}|=H=3$; II. $|\mathcal{S}|=|\mathcal{A}|=|\mathcal{B}|=H=6$. The exploitability is calculated according to Eq.~\eqref{eq:subopt}, which can be solved with dynamic programming in the tabular cases with known transition and reward functions. Our two proposed algorithms without function approximation show significant speedup for decreasing the exploitability compared against other baselines, especially for the larger environment (II). This aligns with our theoretical justification as in Section \ref{sec:theo_just}.

\textbf{With neural networks as function approximation.}
In this set of experimentation, we add neural networks as function approximators. We evaluate \nd and \nde on the same tabular MG environments I and II. We setup the neural-network versions of baseline methods---FSP, NFSP and PSRO, using the same set of hyperparameters (Appendix~\ref{app:hyperparam_tabular}) and the same training configurations. During training, the model checkpoints are saved at different stages for each method and reloaded for exploitation test. Each method is trained for a total of $5\times 10^4$ episodes to get the final model, against which the exploiter is trained for $3\times 10^4$ episodes as the exploitation test. The exploiter is a DQN agent trained from scratch with the same hyperparameters as the base agents (for finding best responses) in FSP, NFSP, PSRO.
We empirically measure the exploitability of the learner's policy as follows: we first compute smoothed version of cumulative utility achieved by the exploiter at each episode (the smoothing is conducted by averaging over a small set of neighboring episodes); we then report the highest smoothed cumulative utility of the exploiter as an approximation for the exploitability.
We also test the effectiveness of using DQN exploiter to measure the exploitability, by training it against the oracle Nash strategies (i.e., the ground-truth Nash equilibria).
Results for two tabular environments are displayed in Table~\ref{tab:tabular_exploit_compare128}. 
As shown in Table~\ref{tab:tabular_exploit_compare128}, both \nd and \nde outperform 
all other methods by a significant margin.
The negative exploitability values of the Oracle Nash strategy indicates that the DQN-based exploiter is not able to find the exact theoretical best response. Nevertheless, it approximates the best response very well. The reported exploitability of Oracle Nash is very close to zero, which justifies the effectiveness of using DQN for exploitation tests. Complete results and details about exploitability calculation are provided in Appendix~\ref{sec:tabular_more_results}.


\begin{table*}[htbp]
\centering
\caption{Exploiter rewards in tabular environments: the reported number is computed over 1000 episodes in exploitation test. }
\resizebox{\textwidth}{!}{ 
\begin{tabular}{ c|c|c|c|c|c|c|c||c } 
\hline
\backslashbox{Env}{Method} & SP & FSP & NFSP & PSRO & Nash DQN & Nash DQN Exploiter & Oracle Nash & Nash V\\
\hline
Tabular Env I & $0.74\pm{0.71}$ & $0.67\pm{0.81}$ & $0.61\pm{0.85}$ & $0.43\pm{1.11}$ & $0.39\pm{1.04}$ & $\color{blue}\mathbf{0.31\pm{0.99}}$ & $0.26\pm{1.02}$ & $-0.29$ \\ \hline
Tabular Env II & $1.37\pm1.32$ & $0.825\pm 1.34$ & $0.51\pm1.40$ & $0.70\pm1.33$ & $\color{blue}\mathbf{0.14\pm1.34}$ & $ 0.20\pm1.42$ & $0.04\pm1.36$ & $-0.13$\\
\hline
\end{tabular}
}
\label{tab:tabular_exploit_compare128}
\end{table*}

\subsection{Two-Player Video Game}
\label{sec:video_game}
To evaluate the scalability and robustness of the proposed method, we examine all algorithms in five two-player Atari environments~\citep{Bellemare_2013} in PettingZoo library~\citep{pettingzoo} (\textit{Boxing-v1}, \textit{Double Dunk-v2}, \textit{Pong-v2}, \textit{Tennis-v2}\footnote{The original \textit{Tennis-v2} environment in PettingZoo is not zero-sum, a reward wrapper is applied to make it.}, \textit{Surround-v1}) and in environment \textit{SlimeVolley-v0} in a public available benchmark named Slime Volley-Ball~\citep{slimevolleygym}, as shown in Fig.~\ref{fig:games}. The algorithms tested for this setting include: (1) SP, (2) FSP, (3) NFSP, (4) PSRO, (5) \nd and (6) \nde. To speed up the experiments, each environment is truncated to 300 steps per episode for both training and exploitation. A full length experiment is conducted on one environment in Appendix.~\ref{app:sec_full_len}. For Atari games, the observation is based on RAM and normalized in range $[0,1]$.

Similar to experiments in the tabular environment with function approximation, the exploitation test (using single-agent DQN) is conducted to evaluate the learned models. 
Ideally, if an agent learns the perfect Nash equilibrium strategy, then by definition, we shall expect the agent to be perfectly non-exploitable (i.e., with even the strongest exploiter only capable of achieving her cumulative utility at most zero in symmetric games).  To carry out the experiment,  we first trained all the methods for $5\times 10^4$ episodes, with detailed hyperparameters provided in Appendix~\ref{app:hyperparam_nontabular}. After the methods are fully trained, we take their final models or certain distributions of historical strategies (uniform for FSP and meta-Nash for PSRO), and the train separate exploiters playing against those learned strategies.
We instantiate a DQN agent as exploiter using the same set of hyperparameters and network architectures to learn from scratch against the fixed trained checkpoint. The resulting learning curve in the exploitation test illustrates the degree of exploitation. An exploiter reward greater than zero indicates that the agent has been exploited since the games are symmetric. The model with lower exploiter reward means is more difficult to be exploited (is thus better).

\begin{figure}[htp]
    \centering
    \includegraphics[width=\columnwidth]{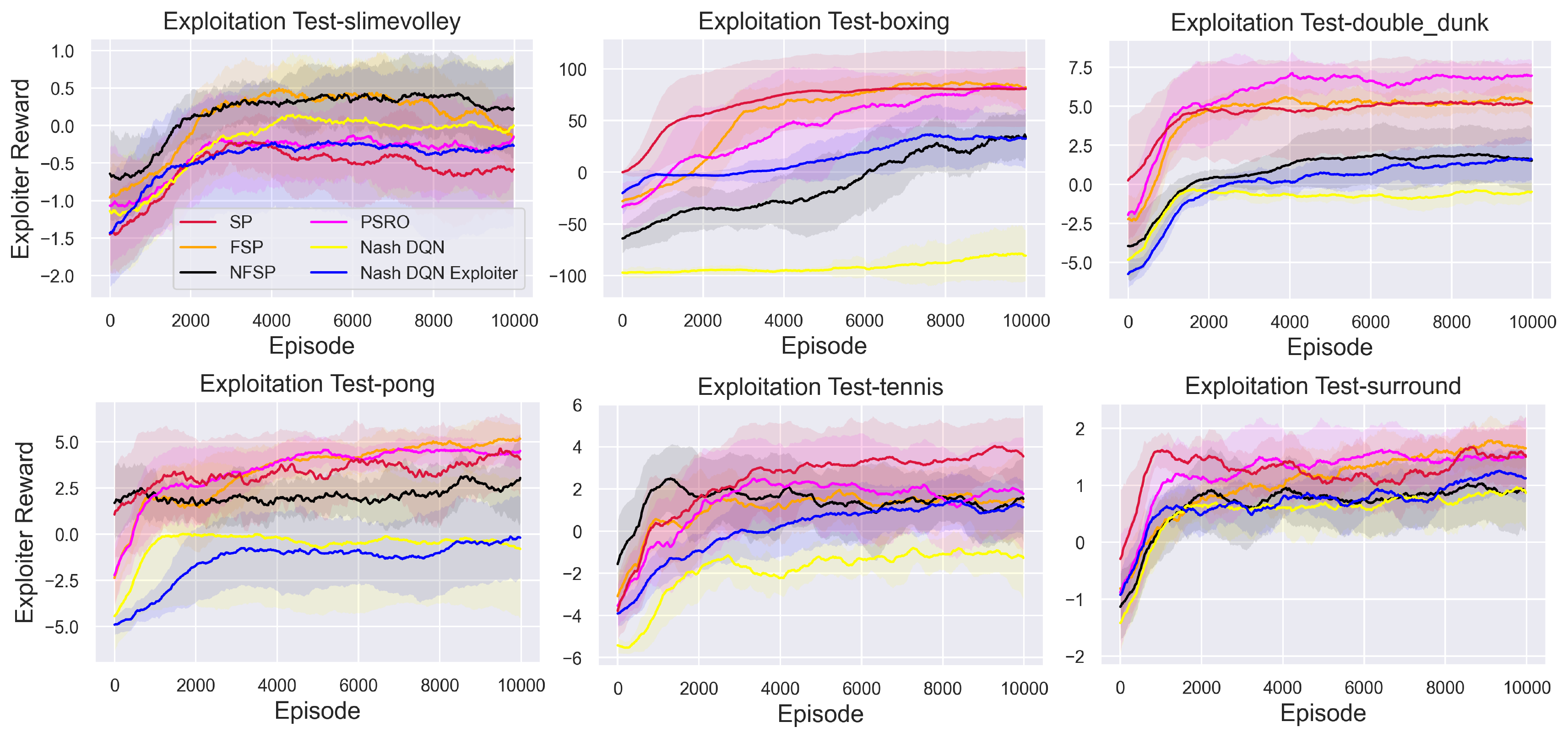}
    \caption{The exploiter learning curves for exploitation tests on six two-player zero-sum video games.}
    \label{fig:atari_nvi}
\end{figure}

\begin{table*}[t]
\scriptsize
\centering
\caption{Approximate exploitability (lower is better) for six two-player video games: mean$\pm$std.}
\vskip -.05in
\resizebox{\textwidth}{!}{ 
\begin{tabular}{>{\centering\arraybackslash}m{50pt}|>{\centering\arraybackslash}m{51pt}|>{\centering\arraybackslash}m{51pt}|>{\centering\arraybackslash}m{51pt}|>{\centering\arraybackslash}m{51pt}|>{\centering\arraybackslash}m{62pt}|>{\centering\arraybackslash}m{51pt}}
\toprule
\multicolumn{1}{c|}{\backslashbox{Env}{Method}} & 
 \multicolumn{1}{c|}{SP}  & \multicolumn{1}{c|}{FSP} & \multicolumn{1}{c|}{NFSP} & \multicolumn{1}{c|}{PSRO} &  \multicolumn{1}{c|}{\textbf{Nash DQN}} & \multicolumn{1}{c}{\textbf{Nash DQN Exploiter}}\\
\hline
  SlimeVolley & $0.00\pm{0.20}$ & $0.54\pm{0.29}$ & $0.58\pm{0.24}$ & $\color{blue}\mathbf{-0.03\pm{0.48}}$ & \cellcolor{gray!15}$0.20\pm{0.71}$ &\cellcolor{gray!15} $-0.02\pm{ 0.87}$\\ \hline
 Boxing & $68.89\pm{42.27}$ & $89.30\pm{9.26}$ & $27.63\pm{26.04}$ & $85.20\pm{15.58}$ &\cellcolor{gray!15} $\color{blue}\mathbf{-74.57\pm{26.96}}$ & $36.92\pm{29.91}$\\  \hline
 Double Dunk & $5.75\pm{2.54}$ & $6.03\pm{ 0.58}$ & $ 2.61\pm{ 1.84}$ & $6.95\pm{0.40}$ & \cellcolor{gray!15}$\color{blue}\mathbf{-0.25\pm{0.42}}$ & $ 1.76\pm{ 1.10}$ \\  \hline
 Pong & $ 4.84\pm{1.23}$ & $ 5.34\pm{0.89}$ & $3.59\pm{0.78}$ & $5.02\pm{0.33}$ & $\cellcolor{gray!15}\color{blue}\mathbf{-1.72\pm{2.61}}$ & $\cellcolor{gray!15} 0.43\pm{2.08}$ \\ \hline
 Tennis  & $4.35\pm{0.87}$ & $ 2.87\pm{ 1.15}$ & $ 3.11\pm{1.09}$ & $3.48\pm{1.74}$ & $\cellcolor{gray!15}\color{blue}\mathbf{-0.35\pm{0.51}}$ & $1.88\pm{1.18}$ \\  \hline
 Surround  & $ 1.64\pm{0.42}$ & $1.77\pm{0.25}$ & $ 1.32\pm{0.49}$ & $1.71\pm{0.40}$ &\cellcolor{gray!15} $\color{blue}\mathbf{1.03\pm{0.29}}$ & $1.33\pm{0.24}$ \\ 
\bottomrule
\end{tabular}
}
\label{tab:atari_exploit}
\end{table*}

Fig.~\ref{fig:atari_nvi} and Table~\ref{tab:atari_exploit} show the exploitation results of all algorithms and baselines. Each method is trained for five random seeds, and the model for each random seed is further exploited with DQN exploiter for $10^4$ episodes under three random initializations. 
For each method and environment, Table~\ref{tab:atari_exploit} displays the best performing models with its corresponding exploitability. Complete results are provided in Appendix Sec.~\ref{app:sec_results_video}. The values in the Table~\ref{tab:atari_exploit} is the maximum of smoothed exploiter reward in the exploitation test of Fig.~\ref{fig:atari_nvi}. The baseline methods SP, FSP, NFSP, PSRO do not perform well in most games. This shows the challenge for finding approximate Nash equilibrium strategies for these games. \nd demonstrates significant advantages over other methods across all six games. Except for \textit{Surround} environment, \nd achieves non-positive exploiter rewards for five environments, which demonstrates the non-exploitability of the policies learned by \nd.
\nde also shows unexploitable performance on \textit{SlimeVolley} and \textit{Pong} environments. Different environments show different levels of difficulties to find a non-exploitable model. \textit{SlimeVolley} is relatively easy with almost all methods achieving exploitability close to zero. \textit{Surround} is generally hard to resolve due to the inherent complexity of the game. We believe that solving Surround requires more advanced exploration technique to boost its performance. 

\begin{wrapfigure}{r}{0.6\textwidth}
\vskip -.15in
  \begin{center}    \includegraphics[width=0.6\columnwidth]{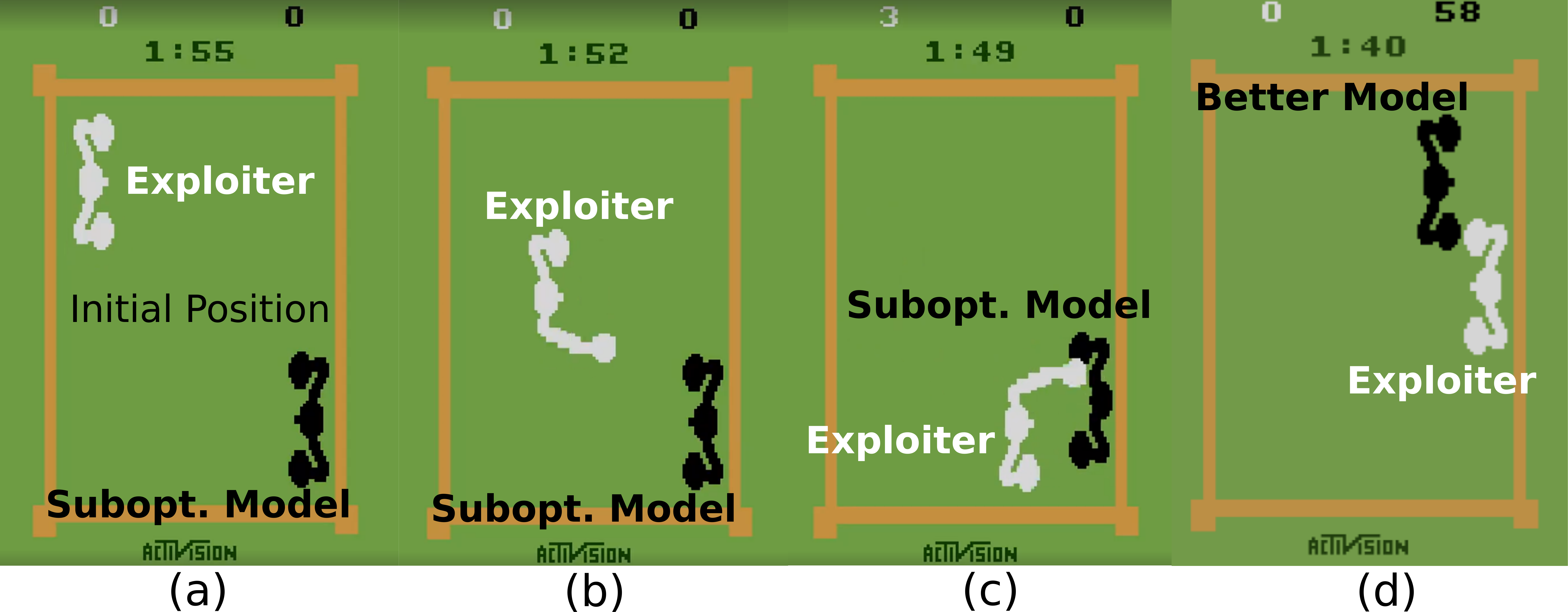}
    \caption{The key frames in \textit{Boxing} exploitation test: (a-c) shows a sub-optimal model exploited by the exploiter. (d) shows our proposed algorithms learn hard-to-exploit policy robust against the turning-around strategy of the exploiter.}
    \label{fig:exploit_box}
  \end{center}
  \vskip -.15in
\end{wrapfigure}

Interestingly, we observe that in the exploitation test for \textit{Boxing} environment, baseline methods such as SP sometimes produce a policy that keeps staying at the corner of the ground. As shown in Fig.~\ref{fig:exploit_box}, the black agent uses the learned suboptimal model by SP algorithm, which tries to avoid any touch with the white opponent (a-b). Such policy (always hide in a corner) is not bad when playing against average player or AI whose policies may not have considered this extreme cases and thus unable to even locate the black agent. However, this policy is very vulnerable to exploitation. Once the exploiter explores the way to touch the black agent, (c) our exploiter learns to heavily exploit such policy in a short time. On the contrary, our algorithm \nd and \nde  will never learn such easy-to-exploit policies. The models learned with \nd and \nde are usually aggressively approaching the exploiter and directly fighting against it, which is found to be harder to exploit in this game. Moreover, our policies are further robust to a turn-around strategy by the exploiter, as shown in Fig.~\ref{fig:exploit_box} (d).

\begin{figure}[htbp]
    \centering
    \includegraphics[width=\columnwidth]{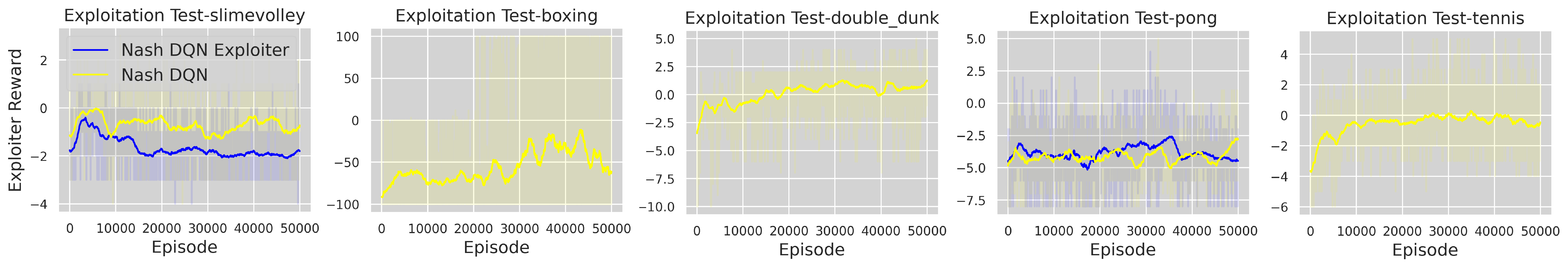}
    \caption{The exploiter learning curves for longer exploitation tests on five video games.}
    \label{fig:exploit_longer}
    \vskip -.15in
\end{figure}

To address the possibility insufficient exploitation on our models, we exploit the models for longer time ($5\times 10^4$ episodes) for those methods within cells shaded in \textcolor{gray}{gray} in Table~\ref{tab:atari_exploit} , and the results are shown in Fig.~\ref{fig:exploit_longer}. Except for the \textit{Double Dunk} environment, the \nd and \nde models are still hard to be exploited on four environments even for long enough exploitation. The difficulty of \textit{Double Dunk} is that each agent needs to control a team of two players to compete through team collaboration, which might require a longer training time to further improve the learned policies.

\section{Conclusion and Discussion}
In this paper, we propose two novel algorithms for deep reinforcement learning in the two-player zero-sum  games. The idea is to incorporate Nash equilibria computation into the training process so that the agents can learn non-exploitable strategies. The first method that we propose is \nd, with the two players trained to learn the Nash value function. To guide its exploration, we also propose a variant of it named \nde, with the learning agent trained to play against an opponent that's designed to exploit learner's weakness. Experimental results demonstrate that the proposed methods converges significantly faster than existing methods in random generated tabular Markov games with or without function approximation. Our experiment further shows that our algorithms are scalable to learn the non-exploitable strategies in most of the six two-player video games, and significantly outperform all the baseline algorithms. This is to our knowledge the first work for learning  non-exploitable strategies in two-player zero-sum video games like Atari.
Limitations also exist for the present methods, the NE solving subroutine is repeatedly called in both inference and update procedures, which leads to large computational costs. Image-based solutions with explicit policies are to be explored in the future.

\bibliographystyle{abbrv}
\bibliography{ref}

\section*{Checklist}


\begin{enumerate}

\item For all authors...
\begin{enumerate}
  \item Do the main claims made in the abstract and introduction accurately reflect the paper's contributions and scope?
    \answerYes{}
  \item Did you describe the limitations of your work?
    \answerYes{}
  \item Did you discuss any potential negative societal impacts of your work?
    \answerNo{To our knowledge, the work has no potential negative societal impacts.}
  \item Have you read the ethics review guidelines and ensured that your paper conforms to them?
    \answerYes{}
\end{enumerate}

\item If you are including theoretical results...
\begin{enumerate}
  \item Did you state the full set of assumptions of all theoretical results?
    \answerYes{}
        \item Did you include complete proofs of all theoretical results?
    \answerNo{} The theoretical results mentioned in the paper is not part of the original contribution of the paper.
\end{enumerate}

\item If you ran experiments...
\begin{enumerate}
  \item Did you include the code, data, and instructions needed to reproduce the main experimental results (either in the supplemental material or as a URL)?
    \answerNo{} The link to the code will be available once the blind review process is passed. The code will be open-sourced with detailed instructions.
  \item Did you specify all the training details (e.g., data splits, hyperparameters, how they were chosen)?
    \answerYes{See Appendix.}
        \item Did you report error bars (e.g., with respect to the random seed after running experiments multiple times)?
    \answerNo{The error bars in our experiments do not reflect the statistical uncertainty due to the inherent randomness in the evaluation process.}
        \item Did you include the total amount of compute and the type of resources used (e.g., type of GPUs, internal cluster, or cloud provider)?
    \answerYes{See Appendix.}
\end{enumerate}

\item If you are using existing assets (e.g., code, data, models) or curating/releasing new assets...
\begin{enumerate}
  \item If your work uses existing assets, did you cite the creators?
    \answerYes{The test environments used are all cited.}
  \item Did you mention the license of the assets?
    \answerNo{The environments are open-sourced.}
  \item Did you include any new assets either in the supplemental material or as a URL?
    \answerNA{}
  \item Did you discuss whether and how consent was obtained from people whose data you're using/curating?
    \answerNA{}
  \item Did you discuss whether the data you are using/curating contains personally identifiable information or offensive content?
    \answerNA{}
\end{enumerate}

\item If you used crowdsourcing or conducted research with human subjects...
\begin{enumerate}
  \item Did you include the full text of instructions given to participants and screenshots, if applicable?
    \answerNA
  \item Did you describe any potential participant risks, with links to Institutional Review Board (IRB) approvals, if applicable?
    \answerNA
  \item Did you include the estimated hourly wage paid to participants and the total amount spent on participant compensation?
    \answerNA
\end{enumerate}

\end{enumerate}


\clearpage
\appendix

\newpage
\section*{Appendix}

\section{Comparison of Nash Solvers for Normal-Form Game}
\label{sec:choose_solver}
In this section we show the Nash equilibrium solvers (\emph{i.e.}, Nash solving subroutine $\textsc{Nash}$) for zero-sum normal-form games, which is an important subroutine for the proposed algorithms. Since NE for zero-sum normal-form games can be solved by linear programming, some packages involving linear programming functions like ECOS, PuLP and Scipy can be leveraged. We also implement a solver based on an iterative algorithm--multiplicative weights update (MWU), which is detailed in Appendix Sec.~\ref{sec:mwu}.

\subsection{Multiplicative Weights Update}
\label{sec:mwu}
The MWU algorithm~\cite{bailey2018multiplicative, daskalakis2018last} is no-regret in online learning setting, which can be used for solving NE in two-player zero-sum normal-form games. Given a payoff matrix $A$ (from the max-player's perspective), the NE strategies will be solved by iteratively applying MWU. Specifically, for $n$-th iteration, state $s$ and actions $(a_i, b_j)$ for the max-player $\mu$ and min-player $\nu$ respectively, $i, j$ are the entry indices of discrete action space, then the update rule of the action probabilities for two players are:
\begin{align}
    \mu^{(n+1)}(a_i|s) = \mu^{(n)}(a_i|s)\frac{e^{\eta (A \boldsymbol{\nu}^{(n)\intercal}(s))_i}}{\sum_{i^\prime} \mu^{(n)}(a_{i^\prime}; s) e^{\eta (A \boldsymbol{\nu}^{(n)\intercal}(s))_{i^\prime}}} \\
     \nu^{(n+1)}(b_j|s) = \nu^{(n)}(b_j|s)\frac{e^{-\eta (A \boldsymbol{\mu}^{(n)\intercal}(s))_j}}{\sum_{j^\prime} \nu^{(n)}(b_{j^\prime}; s) e^{-\eta (A \boldsymbol{\mu}^{(n)\intercal}(s))_{j^\prime}}}    
\end{align}
where $\eta$ is the learning rate.
By iteratively updating each action entry of the strategies with respect to the payoff matrix, MWU is provably converging to NE.

\subsection{Comparison}
\begin{table}[htbp]
\caption{Comparison of different Nash solvers for zero-sum matrix game ($6\times 6$ random matrices).}
\label{tab:compare_ne_solver}
\begin{threeparttable}
\begin{tabular}{cccc}
\toprule
                           & \multirow{1}{*}{Solver} & \multicolumn{1}{c}{Time per Sample (s)} & \multicolumn{1}{c}{Solvability} \\ \cline{2-4} 
                          
                           & Nashpy (equilibria)\tnote{1}                   & $0.751$  & all \\ \cline{2-4} 
                           & Nashpy (equilibrium)                     & $0.0016$  & not for some degenerate matrices\\ \cline{2-4}
                           & ECOS\tnote{2}                       & $0.0015$                & all \\ \cline{2-4} 
                           & MWU (single)            & $0.008$              & all (but less accurate) \\ \cline{2-4} 
                           & MWU (parallel)                    & fast, depends on batch size\tnote{3}                & all (but less accurate)   \\ \cline{2-4}
                           & CVXPY                  & $0.009$               & all  \\ \cline{2-4} 
                           & PuLP                   & $0.020$             & all \\ \cline{2-4} 
                           & Scipy               & $-$                & not for some \\ \cline{2-4} 
                           & Gurobipy            & $~0.01$    & not for some \\ \midrule
\end{tabular}
\begin{tablenotes}\footnotesize
\item[1] Nashpy (\href{https://github.com/drvinceknight/Nashpy}{https://github.com/drvinceknight/Nashpy}) can be adopted to achieve two versions of NE solvers: one returns a single NE, and another returns all Nash equilibria for given payoff matrices.
\item[2] ECOS (\href{https://github.com/embotech/ecos}{https://github.com/embotech/ecos}) is a package for solving convex second-order cone programs.
\item[3] MWU solver is self-implemented for solving either a single payoff matrix or solving a batch of matrices in parallel.
\item[4] CVXPY (\href{https://github.com/cvxpy/cvxpy}{https://github.com/cvxpy/cvxpy}) is a Python package for convex optimization.
\item[5] PuLP (\href{https://github.com/coin-or/pulp}{https://github.com/coin-or/pulp}) is a linear programming package with Python.
\item[6] Scipy (\textit{scipy.optimize.linprog()}) is a general package for numerical operations.
\item[7] Gurobipy (\href{https://www.gurobi.com/}{https://www.gurobi.com/}) is a package for linear and quadratic optimization.
\end{tablenotes}
\end{threeparttable}
\end{table}  

We conduct experiments to compare different solvers for NE subroutine, including Nashpy (for single Nash \reb{equilibrium} or all Nash equilibria), ECOS, MWU (single or parallel), CVXPY, PuLP, Scipy, Gurobipy.
The test is conducted on a Dell XPS 15 laptop with only CPU computation. The code will be released after the review process (anonymous during review process). Experiments are evaluated on $6\times 6$ random matrices and averaged over 1000 samples. The zero-sum property of the generated Markov games is guaranteed by generating each random matrix as one player's payoff and take the negative values as its opponent's payoff.

As in Table.~\ref{tab:compare_ne_solver}, the solvability indicates whether the solver can solve all possible randomly generated matrices (zero-sum). Nashpy for solving all equilibria cannot handle some degenerate matrices. Scipy and Gurobipy also cannot solve for some payoff matrices. Other solvers can solve all random payoff matrices in our tests but with different solving speed and accuracy. The solvability is essential for the program since the values within the payoff matrix can be arbitrary as a result of applying function approximation. The results support our choice of using the ECOS-based solver as the default Nash solving subroutine for the proposed algorithms, due to its speed and robustness for solving random matrices. ECOS is originally built for solving convex second-order cone programs, which covers linear programming (LP) problem. It tries to transform the input matrices to be Scipy sparse matrices and speeds up the solving procedure. By formulating the NE solving as LP on normal-form game, we can plug in the ECOS solver to get the solution. Some constraints like positiveness and constant sum need to be handled carefully. Other solvers like Nashpy (equilibrium) and MWU (parallel) can achieve a similar level of computational time, but less preferred due to either not being able to solve some matrices or less accurate results. Specifically,  for the case with a large batch size and a small number of inner-loop iterations for MWU, MWU can be faster than ECOS. However, the accuracy of MWU depends on the number of iterations for update~\cite{bailey2018multiplicative, daskalakis2018last}. More iterations lead to more accurate approximation but also longer computational time for MWU method. Empirically, we find that the accuracy of the subroutine solver is critical for our proposed algorithms with function approximation, especially in video games with long horizons. \reb{Moreover, although we have already selected the best solver through comparisons, the computational time of the solvers used in each inference or update step still account for a considerable portion. This leaves some space for improvement of running-time efficiency. }


\section{Algorithms on Tabular Markov Games}
\label{sec:alg_tab_markov}
In this section, we provide further details on connections of our algorithms to tabular algorithms \nv and \nve in Sec.\ref{sec:connections}, and prove the theoretical guarantees of the latter two algorithms in Sec.\ref{sec:proof_of_theory}. We then provide all the pseudo-codes for subroutines and algorithms used in this paper. In particular, we introduce the pseudo-codes for important subroutines in Sec.~\ref{app:sec_sub}, and then the pseudo-codes for several algorithms: self-play (SP, \ref{app:sec_sp}), fictitious self-play (FSP, \ref{app:sec_fsp}),  double oracle (DO, \ref{app:sec_do}), Nash value iteration (\nv, \ref{app:sec_nvi}) and Nash value iteration with exploiter (\nve, \ref{app:sec_nvie}). SP, FSP and DO are the baseline methods in experimental comparisons, while \nv and \nve are the tabular version of our proposed algorithms \nd and \nde, without function approximation.


\subsection{Connections of \nd, \nde to tabular algorithms} \label{sec:connections}
We first note that the $\epsilon$-greedy version of \nv and \nve algorithms (as shown in Section \ref{app:sec_nvi}, \ref{app:sec_nvie}), are simply the optimistic Nash-VI algorithm in \cite{liu2021sharp} and \golf algorithm in \cite{jin2021power} when applied to the tabular setting, with optimistic exploration replaced by $\epsilon$-greedy exploitation.

Comparing our algorithms \nd (algorithm \ref{alg:nash_dqn}) and \nde (algorithm \ref{alg:nash_dqn_exp}) with the $\epsilon$-greedy version of \nv (Algorithm \ref{alg:sec_nvi}) and \nve (Algorithm \ref{alg:sec_nvie}), we notice that, besides the minor difference between episodic setting versus infinite horizon discounted setting, the latter two algorithms are special cases of the former two algorithms when
\begin{enumerate}
    \item specialize the neural network structure to represent a table of values for each state-action pairs (i.e. specialize both algorithms to the tabular setting);
    \item let the minibatch $\mathcal{M}$ to contain all previous data (i.e., use the full batch $\cD$);
    \item let the number of gradient step $m$ to be sufficiently large so that GD finds the minimizer of the objective function;
    \item let $N=1$, that is update the target network at every iterations.
\end{enumerate}

We remark that the use of small minibatch size, and small gradient steps are to speed up the training in practice beyond tabular settings. The delay update of the target networks is used to stabilize the training process.

\subsection{Proof of Theorem \ref{thm:theory}} \label{sec:proof_of_theory}
The result of optimistic Nash-VI algorithm in \cite{liu2021sharp}, and the result of \golf algorithm in \cite{jin2021power} (when specialized to the tabular setting) already prove that both optimistic versions of \nv and \nve can find
$\epsilon$-approximate Nash equilibria for \textbf{episodic} Markov games in $\text{poly}(S, A, B, H, \epsilon^{-1}, \log(1/\delta))$ steps with probability at least $1-\delta$. Here $H$ is the horizon length of the episodic Markov games.

To convert the episodic results to the infinite-horizon discounted setting in this paper, we can simply truncate the infinite-horizon games up to $H = \frac{1}{1-\gamma} \ln \frac{2}{(1-\gamma)\epsilon}$ steps so that the remaining cumulative reward is at most
\begin{equation*}
    \sum_{h=H}^\infty \gamma^h = \frac{\gamma^H}{1-\gamma} \le \frac{e^{-(1-\gamma)H}}{1-\gamma} \le \frac{\epsilon}{2}
\end{equation*}
which is smaller than the target accuracy.  To further address the non-stationarity of the value and policy in the the episodic setting (which requires both value and policy to depends on not only the state, but also the steps), we can augment the state space $s$ to $(s, h)$ to include step information (up to the truncation point $H$) in the state space. Now, we are ready to apply the episodic results to the infinite horizon discounted setting, which shows that both optimistic versions of \nv and \nve can find
$\epsilon$-approximate Nash equilibria for \textbf{infinite-horizon discounted} Markov games in $\text{poly}(S, A, B, (1-\gamma)^{-1}, \epsilon^{-1}, \log(1/\delta))$ steps with probability at least $1-\delta$. Here $\gamma$ is the discount coefficient.

\subsection{Subroutines}
\label{app:sec_sub}

\begin{algorithm}[htbp]
\caption{\textcolor{magenta}{$\textsc{Meta\_Nash}$}: Meta-Nash Equilibrium Solving Subroutine}
\begin{algorithmic}[1]
\INPUT two strategy sets $\mu,\nu$; evaluation iterations $N$
\STATE Initialize payoff matrix: $M_{i,j}=0, i\in[|\mu|], j\in[|\nu|]$
\FOR{$\mu_i\in \mu$}
\FOR{$\nu_i\in \nu$}
\FOR{episodes $k=1,\ldots,N$}
\STATE Rollout policies $\mu_i, \nu_j$ to get episodic reward $r_k$
\STATE $M_{i,j}=\frac{1}{N}\sum_{k=1}^N r_k$
\ENDFOR
\ENDFOR
\ENDFOR
\STATE $(\rho_\mu, \rho_\nu)= \textcolor{magenta}{\textsc{Nash}}(M)$ 
\STATE Return $\rho_\mu$ or $\rho_\nu$
\end{algorithmic}
\label{alg:meta_nash}
\end{algorithm}

Before introducing the pseudo-code for each algorithm, we summarize several subroutines -- {\color{magenta} $\textsc{Nash}$}, {\color{magenta} $\textsc{Meta\_Nash}$}, {\color{magenta} $\textsc{Best\_Response}$} and {\color{magenta} $\textsc{Best\_Response\_Value}$} -- applied in the algorithms. These subroutines are marked in \textcolor{magenta}{magenta} color in the this and the following sections.

{\color{magenta} $\textsc{Nash}$}: As a NE solving subroutine for normal-form games, it returns the NE strategy given the payoff matrix as the input. Specifically it uses the solvers introduced in Appendix Sec.~\ref{sec:choose_solver}, and ECOS is the default choice in our experiments.

{\color{magenta} $\textsc{Meta\_Nash}$}: As a meta-Nash solving subroutine (Algorithm~\ref{alg:meta_nash}), it returns the one-side meta NE strategy given two strategy \textbf{sets}: $\mu=\{\mu_1, \cdots, \mu_i, \cdots\},\nu=\{\nu_1, \cdots, \nu_i, \cdots\}$. A one-by-one matching for each pair of polices $(\mu_i, \nu_j), i\in [|\mu|], j\in[|\nu|]$ is evaluated in the game to get an estimated payoff matrix, with the average episodic return as the estimated payoff values of two players for each entry in the payoff matrix. The {\color{magenta} $\textsc{Nash}$} subroutine is called to solve the meta-Nash strategies. It is applied in DO algorithm, which is detailed in Sec.~\ref{app:sec_do}.

\begin{algorithm}[htbp]
\caption{\textcolor{magenta}{$\textsc{Best\_Response}$ I}: Best Response Subroutine in Markov Game (known transition, reward functions)}
\label{alg:br1}
\begin{algorithmic}[1]
\INPUT mixture policy $\rho_\mu$ as a distribution over $\{\mu^0, \mu^1,\dots, \mu^n\}$
\STATE Initialize non-Markovian policies $\hat{\mu}=\{\hat{\mu}_h\}, \hat{\nu}=\{\hat{\nu}_h\}, h\in[H], \mu_h:(\mathcal{S}\times\mathcal{A\times\mathcal{B}})^{(h-1)}\times\mathcal{S}\times\mathcal{A}\rightarrow [0,1], \nu_h:(\mathcal{S}\times\mathcal{A\times\mathcal{B}})^{(h-1)}\times\mathcal{S}\times\mathcal{B}\rightarrow [0,1]$
\STATE Initialize $Q$ table for non-Markovian policies $\hat{\mu}, \hat{\nu}$, $Q=\{Q_h\}, h\in[H], Q_h:(\mathcal{S}\times\mathcal{A}\times\mathcal{B})^{h}\rightarrow [0,1]$
\STATE Initialize $V$ table for non-Markovian policies $\hat{\mu}, \hat{\nu}$, $V=\{V_h\}, h\in[H], V_h:(\mathcal{S}\times\mathcal{A}\times\mathcal{B})^{(h-1)}\times\mathcal{S}\rightarrow [0,1]$
\FOR{$h=1,\ldots,H$}
\STATE For all $\tau_{h-1}$:
\begin{align}
    Q^{\mu, \dagger}_h(\tau_{h-1}, s_h,a_h,b_h) &=\sum_{s^\prime\in\mathcal{S}}\mathbb{P}_h(s_{h+1}|s_h,a_h,b_h)[r_h(s_h, a_h, b_h)+V^{\hat{\mu}, \dagger}_{h+1}(\tau_h, s_{h+1})] \\
    V^{\hat{\mu}, \dagger}_h(\tau_{h-1}, s_h) &= \min_{\nu_h}\hat{\mu}_h(\cdot|\tau_{h-1}, s_h) Q_h^{\hat{\mu}, \dagger}(\tau_{h-1}, s_h, \cdot, \cdot)\nu_h^\intercal(\cdot|\tau_{h-1}, s_h)\\
    \hat{\nu}_h(\tau_{h-1}, s_h) &= \arg\min_{\nu_h}\hat{\mu}_h(\cdot|\tau_{h-1}, s_h) Q_h^{\hat{\mu}, \dagger}(\tau_{h-1}, s_h, \cdot, \cdot)\nu_h^\intercal(\cdot|\tau_{h-1}, s_h)
\end{align}
where
\begin{equation*}
    \hat{\mu}_h(a_h|\tau_{h-1}, s_h):=\frac{\sum_i \mu_h^i(a_h| s_h)\rho(i)\Pi_{t^\prime=1}^{h-1}\mu^i_{t^\prime}(a_{t^\prime}| s_{t^\prime} )}{\sum_j \rho(j)\Pi_{t^\prime=1}^{h-1}\mu^j_{t^\prime}(a_{t^\prime}| s_{t^\prime} )}
\end{equation*}
\ENDFOR
\STATE Return $\hat{\nu}$ or $V_1^{\hat{\mu}, \dagger}(s_1)$ \\
{\color{blue}\% $\hat{\mu}$ is the posterior policy of non-Markovian mixture $\mu$, $\hat{\nu}$ is the best response of it}
\end{algorithmic}
\end{algorithm}

\begin{algorithm}[htbp]
\caption{\textcolor{magenta}{$\textsc{Best\_Response}$ II}: Best Response Subroutine in Markov Game ($Q$-learning based, unknown transition, reward functions)}
\label{alg:br2}
\begin{algorithmic}[1]
\INPUT mixture policy $\rho_\mu$ as a distribution over $\{\mu^0, \mu^1,\dots, \mu^n\}$; best response $Q$-learning iterations $N$; soft update coefficient $\alpha$
\STATE Initialize the $Q=\{Q_h|h\in[H]\}\in \mathbb{R}^{|\mathcal{S}| \times |\mathcal{B}|}$ table for the best response player, 
\FOR{episodes $k=1,\ldots,N$}
\STATE Sample policy $\mu_k\sim\rho_\mu$ 
\FOR{$t = 1,\ldots,H$}
\STATE {\color{blue}\% collect data}
\STATE Sample greedy action $a_t\sim \mu_k(\cdot|s_t)$
\STATE With $\epsilon$ probability, sample random action $b_t$;
\STATE Otherwise, sample greedy action $b_t\sim\nu(\cdot|s_t)$ according to $Q$ 
\STATE Rollout environment to get sample $(s_t,a_t,b_t,r_t,\text{done},s_{t+1})$ \quad($r_t$ is for the learning player)
\STATE {\color{blue}\% update best response Q-value}
\IF{not done} \STATE $Q^\text{target}_t(s_t, b_t)=r_t+ V_{t+1}(s_{t+1})$
\STATE where $V_{t+1}(s_{t+1})=\max_{b'}Q_{t+1}(s_{t+1}, b')$
\ELSE \STATE $Q_t^\text{target}(s_t, b_t)=r_t$
\ENDIF
\STATE $Q_t(s_t, b_t)\leftarrow \alpha\cdot Q_t^\text{target}(s_t, b_t)+(1-\alpha)\cdot Q_t(s_t, b_t)$
\IF{done} \STATE break
\ENDIF
\ENDFOR
\ENDFOR
\STATE Represent $Q$ as a greedy policy $\hat{\nu}$
\STATE Return $\hat{\nu}$
\end{algorithmic}
\end{algorithm}

{\color{magenta} $\textsc{Best\_Response}$}: As a best response subroutine, it returns the best response strategy of the given strategy, which satisfies Eq.~\eqref{eq:best_response_v}. To be noticed, the best response we discuss here is the best response of a meta-distribution $\rho_\mu$ over a strategy set $\{\mu^0, \mu^1,\dots, \mu^n\}$, which covers the case of best response against a single strategy by just making the distribution one-hot. We use this setting for the convenience of being applied in SP, FSP, DO algorithms. Here we discuss two types of best response subroutine that are used at different positions in the algorithms: (1) \textcolor{magenta}{$\textsc{Best\_Response}$ I} (as Algorithm~\ref{alg:br1}) is a best response subroutine with oracle transition and reward function of the game, which is used for evaluating the exploitability of the model after training; (2) \textcolor{magenta}{$\textsc{Best\_Response}$ II} (as Algorithm~\ref{alg:br2}) is a best response subroutine with $Q$-learning agent for approximating the best response, without knowing the true transition and reward function of the game. It is used in the procedure of methods based on iterative best response, like SP, FSP, DO. We claim here for the following sections, by default, \textcolor{magenta}{ $\textsc{Best\_Response}$} will use \textcolor{magenta}{$\textsc{Best\_Response}$ II}, and \textcolor{magenta}{ $\textsc{Best\_Response\_Value}$} will use \textcolor{magenta}{$\textsc{Best\_Response}$ I}.

{\color{magenta} $\textsc{Best\_Response\_Value}$}: It has the same procedure as $\textcolor{magenta}{\textsc{Best\_Response}}$ as a best response subroutine, but returns the average value of the initial states as $V_1^{\hat{\mu}, \dagger}(s_1)$ in Eq.~\eqref{eq:subopt} with the given strategy $\hat{\mu}$. Since the best response value estimation is used in evaluating the exploitability of a certain strategy, it by default adopts \textcolor{magenta}{$\textsc{Best\_Response}$ I} (Algorithm~\ref{alg:br1}) as an oracle process, which returns the ground-truth best response values because of knowing the transition and reward functions.

\subsection{\reb{Nash Q-Learning}}
\label{app:sec_nash_q_l}
\reb{
The pseudo-code for Nash Q-Learning is shown in Algorithm.\ref{alg:nash_q_l} below.
\begin{algorithm}[H]
\caption{Nash Q-Learning}
\label{alg:nash_q_l}
\begin{algorithmic}[1]
	\STATE Initialize $Q:\mathcal{S}\times\mathcal{A}\times\mathcal{B}\rightarrow \mathbb{R}$, given $\epsilon, \gamma, \alpha$.
\FOR{$k = 1,\ldots,K$}
\FOR{$t = 1,\ldots,H$}
\STATE {\color{blue}\% collect data}
\STATE With $\epsilon$ probability, sample random actions $a_t, b_t$;
\STATE Otherwise, $a_t\sim\mu(\cdot|s_t),b_t\sim\nu(\cdot|s_t), (\mu(\cdot|s_t),\nu(\cdot|s_t) )=\textcolor{magenta}{\textsc{Nash}}(Q(s_{t}, \cdot, \cdot))$ 
\STATE Rollout environment to get sample $(s_t,a_t,b_t,r_t,\text{done},s_{t+1})$
\STATE {\color{blue}\% update Q-value}
\IF{not done}
\STATE Compute $(\hat{\mu}, \hat{\nu})=\textcolor{magenta}{\textsc{Nash}}(Q(s_{t+1}, \cdot, \cdot))$
\STATE Set $Q^\text{target}(s_t, a_t, b_t) =  r_t + \gamma \hat{\mu}\trans Q(s_{t+1}, \cdot, \cdot)\hat{\nu}$.
\ELSE \STATE Set $Q^\text{target}(s_t, a_t, b_t)=r_t$
\ENDIF
\STATE $Q(s_t, a_t, b_t)\leftarrow \alpha\cdot Q^\text{target}(s_t, a_t, b_t)+(1-\alpha)\cdot Q(s_t, a_t, b_t)$
\IF{done} \STATE break
\ENDIF
\ENDFOR
\ENDFOR
\end{algorithmic}
\end{algorithm}
}

\subsection{Self-play}
\label{app:sec_sp}
The pseudo-code for self-play is shown in Algorithm \ref{alg:sp}.

\begin{algorithm}[htbp]
\caption{Self-play for Markov Game}
\begin{algorithmic}[1]
\STATE Initialize policies $\mu^0=\{\mu_h\}, \nu^0=\{\nu_h\}, h\in[H]$
\STATE Initialize policy sets: $\mu=\{\mu^0\}, \nu=\{\nu^0\}$
\STATE Initialize meta-strategies: $\rho_\mu=[1.], \rho_\nu=[1.]$
\FOR{$t=1,\ldots,T$}
    \IF{$t\%2==0$}
     \STATE  $\nu^t = \textcolor{magenta}{\textsc{Best\_Response}}(\rho_\mu, \mu)$
     \STATE $\nu=\nu\bigcup\{\nu^t\}$
     \STATE $\rho_\nu=(0,\dots, 1)$ as a one-hot vector with only 1 for the last entry
    \ELSE
    \STATE  $\mu^t = \textcolor{magenta}{\textsc{Best\_Response}}(\rho_\nu, \nu)$
    \STATE $\mu=\mu\bigcup\{\mu^t\}$
    \STATE $\rho_\mu=(0,\dots, 1)$ as a one-hot vector with only 1 for the last entry
    \ENDIF
    \STATE exploitability = $\textcolor{magenta}{\textsc{Best\_Response\_Value}}(\rho_\mu,\mu)+\textcolor{magenta}{\textsc{Best\_Response\_Value}}(\rho_\nu,\nu)$
\ENDFOR
\STATE Return $\mu, \nu$
\end{algorithmic}
\label{alg:sp}
\end{algorithm}

\subsection{Fictitious Self-play}
\label{app:sec_fsp}
The pseudo-code for fictitious self-play is shown in Algorithm.\ref{alg:fsp}. We use uniform($\cdot$) to denote a uniform distribution over the policy set.

\begin{algorithm}[htbp]
\caption{Fictitious Self-play for Markov Game}
\label{alg:fsp}
\begin{algorithmic}[1]
\STATE Initialize policies $\mu^0=\{\mu_h\}, \nu^0=\{\nu_h\}, h\in[H]$
\STATE Initialize policy sets: $\mu=\{\mu^0\}, \nu=\{\nu^0\}$
\STATE Initialize meta-strategies: $\rho_\mu=[1.], \rho_\nu=[1.]$

\FOR{$t=1,\ldots,T$}
    \IF{$t\%2==0$}
     \STATE  $\nu^t = \textcolor{magenta}{\textsc{Best\_Response}}(\rho_\mu, \mu)$
     \STATE $\nu=\nu\bigcup\{\nu^t\}$
     \STATE $\rho_\nu=\text{Uniform}(\nu)$
    \ELSE
    \STATE  $\mu^t = \textcolor{magenta}{\textsc{Best\_Response}}(\rho_\nu, \nu)$
    \STATE $\mu=\mu\bigcup\{\mu^t\}$
    \STATE $\rho_\mu=\text{Uniform}(\mu)$
    \ENDIF
    \STATE exploitability = $\textcolor{magenta}{\textsc{Best\_Response\_Value}}(\rho_\mu,\mu)+\textcolor{magenta}{\textsc{Best\_Response\_Value}}(\rho_\nu,\nu)$
\ENDFOR
\STATE Return $\mu, \rho_\mu, \nu, \rho_\nu$
\end{algorithmic}
\end{algorithm}

\begin{algorithm}[htbp]
\caption{Double Oracle for Markov Game}
\label{alg:sec_do}
\begin{algorithmic}[1]
\STATE Initialize policies $\mu^0=\{\mu_h\}, \nu^0=\{\nu_h\}, h\in[H]$
\STATE Initialize policy sets: $\mu=\{\mu^0\}, \nu=\{\nu^0\}$
\STATE Initialize meta-strategies: $\rho_\mu=[1.], \rho_\nu=[1.]$

\FOR{$t=1,\ldots,T$}
    \IF{$t\%2==0$}
     \STATE  $\nu^t=\textcolor{magenta}{\textsc{Best\_Response}}(\rho_\mu, \mu)$
     \STATE $\nu=\nu\bigcup\{\nu^t\}$
     \STATE $\rho_\nu=\textcolor{magenta}{\textsc{Meta\_Nash}}(\nu, \mu)$
    \ELSE
    \STATE  $\mu^t=\textcolor{magenta}{\textsc{Best\_Response}}(\rho_\nu, \nu)$
    \STATE $\mu=\mu\bigcup\{\mu^t\}$
    \STATE $\rho_\mu=\textcolor{magenta}{\textsc{Meta\_Nash}}(\nu, \mu)$
    \ENDIF
    \STATE exploitability = $\textcolor{magenta}{\textsc{Best\_Response\_Value}}(\rho_\mu, \mu)+\textcolor{magenta}{\textsc{Best\_Response\_Value}}(\rho_\nu,\nu)$
\ENDFOR
\STATE Return $\mu, \rho_\mu, \nu, \rho_\nu$
\end{algorithmic}
\end{algorithm}

\subsection{Double Oracle}
\label{app:sec_do}
The pseudo-code for double oracle is shown in Algorithm.\ref{alg:sec_do}.

\subsection{\nv}
\label{app:sec_nvi}
The pseudo-code for Nash value iteration (\nv) is shown in Algorithm~\ref{alg:sec_nvi}. Different from \nd (as Algorithm~\ref{alg:nash_dqn}), for tabular Markov games, the $Q$ network is changed to be the $Q$ table and updated in a tabular manner (as Algorithm~\ref{alg:sec_nvi}  line~\ref{line:update_nash_vi}), given the estimated transition function $\tilde{\mathbb{P}}$ and reward function $\tilde{r}$. The target $Q$ is not used. 
Since \nv is applied for tabular Markov games, here  we write the pseudo-code in an episodic setting without the reward discount factor, which is slightly different from Sec.~\ref{subsec:nash_dqn}.

\begin{algorithm}[htbp]
\caption{Nash Value Iteration
\label{alg:sec_nvi}
(\nv, $\epsilon$-greedy sample version)}
\begin{algorithmic}[1]
	\STATE Initialize $Q=\{Q_h\},h\in[H],Q_h:\mathcal{S}_h\times\mathcal{A}_h\times\mathcal{B}_h\rightarrow \mathbb{R}$, buffer $\cD=\phi$, given $\epsilon$, update interval $p$.
\FOR{$k = 1,\ldots,K$}
\FOR{$t = 1,\ldots,H$}
\STATE {\color{blue}\% collect data}
\STATE With $\epsilon$ probability, sample random actions $a_t, b_t$;
\STATE Otherwise, $a_t\sim\mu_t(\cdot|s_t),b_t\sim\nu_t(\cdot|s_t), (\mu_t(\cdot|s_t),\nu_t(\cdot|s_t) )=\textcolor{magenta}{\textsc{Nash}}(Q_t(s_{t}, \cdot, \cdot))$.
\STATE Rollout environment to get sample $(s_t,a_t,b_t,r_t,\text{done},s_{t+1})$ and store in $\cD$.
\STATE {\color{blue}\% update Q-value}
\IF{$|\cD|\%p=0$}
\FOR {$\forall (s,a,b, s')\in\mathcal{S}_h\times\mathcal{A}_h\times\mathcal{B}_h\times\mathcal{S}_{h+1},  h\in [H]$}
\STATE Estimate $\tilde{\mathbb{P}}_h(s_{h+1}=s'|s_h=s,a_h=a,b_h=b)=\frac{1}{n}\sum_{i=1}^n \mathbbm{1}(s_{h+1}=s'_i), (s,a,b,s_i')\in \cD$.
\STATE Estimate $\tilde{r}_h(s_h=s,a_h=a,b_h=b)=\frac{1}{m}\sum_{i=1}^m r_i(s,a,b), (s,a,b,r_i)\in \cD$.
\STATE \alglinelabel{line:update_nash_vi} $Q_h(s, a, b)=\tilde{r}_h(s,a,b)+ (\tilde{\mathbb{P}}_hV_{h+1}^{\hat{\mu}_{h+1}, \hat{\nu}_{h+1}})(s,a, b)\cdot \mathbb{I}[s' \text{ is non-terminal}]$,

\STATE where $(\hat{\mu}_{h+1}, \hat{\nu}_{h+1})=\textcolor{magenta}{\textsc{Nash}}(Q_{h+1})$.
\ENDFOR
\ENDIF
\IF{done} \STATE break
\ENDIF
\ENDFOR
\ENDFOR
\end{algorithmic}
\end{algorithm}

\subsection{\nve}
\label{app:sec_nvie}
The pseudo-code for Nash value iteration with Exploiter (\nve) is shown in Algorithm.\ref{alg:sec_nvie}.  Different from \nde (as Algorithm~\ref{alg:nash_dqn_exp}), for tabular Markov games, the $Q$ network and exploiter $\tilde{Q}$ network are changed to be $Q$ tables and updated in a tabular manner (as Algorithm~\ref{alg:sec_nvie}  line~\ref{line:update_nash_vi_exp} and line~\ref{line:update_nash_vi_exp_tilde}), given the estimated transition function $\tilde{\mathbb{P}}$ and reward function $\tilde{r}$. The target $Q$ and target $\tilde{Q}$ are not used. Since \nve is applied for tabular Markov games, here  we write the pseudo-code in an episodic setting without the reward discount factor, which is slightly different from Sec.~\ref{subsec:nash_dqn_exploiter}.

\begin{algorithm}[htbp]
\caption{Nash Value Iteration with Exploiter (\nve, $\epsilon$-greedy sample version)}
\label{alg:sec_nvie}
\begin{algorithmic}[1]
	\STATE Initialize $Q=\{Q_h\}, \tilde{Q}=\{\tilde{Q}_h\},h\in[H],\tilde{Q}_h,Q_h:\mathcal{S}_h\times\mathcal{A}_h\times\mathcal{B}_h\rightarrow \mathbb{R}$, buffer $\cD=\phi$, given $\epsilon$, update interval $p$.
\FOR{$k = 1,\ldots,K$}
\FOR{$t = 1,\ldots,H$}
\STATE {\color{blue}\% collect data}
\STATE With $\epsilon$ probability, sample random actions $a_t, b_t$;
\STATE Otherwise, $a_t\sim\mu_t(\cdot|s_t),b_t\sim\tilde{\nu}_t(\cdot|s_t)$, 
\STATE $(\mu_t(\cdot|s_t),\nu_t(\cdot|s_t) )=\textcolor{magenta}{\textsc{Nash}}(Q(s_{t}, \cdot, \cdot)), \tilde{\nu}_t(\cdot|s_t) = \arg\min_\nu  \mu_t^\intercal(\cdot|s_t) \tilde{Q}_t(s_t, \cdot, \cdot)\nu$.
\STATE Rollout environment to get sample $(s_t,a_t,b_t,r_t,\text{done},s_{t+1})$ and store in $\cD$.
\STATE {\color{blue}\% update Q-value}
\IF{$|\mathcal{D}|\%p=0$}
\FOR {$\forall (s,a,b, s')\in\mathcal{S}_h\times\mathcal{A}_h\times\mathcal{B}_h\times\mathcal{S}_{h+1},  h\in [H]$}
\STATE Estimate $\tilde{\mathbb{P}}_h(s_{h+1}=s'|s_h=s,a_h=a,b_h=b)=\frac{1}{n}\sum_{i=1}^n \mathbbm{1}(s_{h+1}=s'_i), (s,a,b,s_i')\in \cD$.
\STATE Estimate $\tilde{r}_h(s_h=s,a_h=a,b_h=b)=\frac{1}{m}\sum_{i=1}^m r_i(s,a,b), (s,a,b,r_i)\in \cD$.
\STATE \alglinelabel{line:update_nash_vi_exp} $Q_h(s, a, b)=\tilde{r}_h(s,a,b)+ (\tilde{\mathbb{P}}_hV_{h+1}^{\hat{\mu}_{h+1}, \hat{\nu}_{h+1}})(s,a, b)\cdot \mathbb{I}[s' \text{ is non-terminal}]$,

\STATE where $(\hat{\mu}_{h+1}, \hat{\nu}_{h+1})=\textcolor{magenta}{\textsc{Nash}}(Q_{h+1})$.
\STATE \alglinelabel{line:update_nash_vi_exp_tilde} $\tilde{Q}_h(s, a, b)=\tilde{r}_h(s,a,b)+ (\tilde{\mathbb{P}}_hV_{h+1}^\text{Exploit})(s,a, b)$,

\STATE where $V_{h+1}^\text{Exploit}(s')=\begin{cases}\min_{b'\in\mathcal{B}_{h+1}}{\hat{\mu}}_{h+1}(s')^\intercal \tilde{Q}_{h+1}(s', \cdot, b') & \text{for non-terminal $s'$}\\0 & \text{for terminal $s'$} \end{cases}$.
\ENDFOR
\ENDIF
\IF{done} \STATE break
\ENDIF
\ENDFOR
\ENDFOR
\end{algorithmic}
\end{algorithm}




\subsection{\reb{Comparisons of \nv, Nash Q-Learning, \golf and \nd}}
\label{app:sec_compare_tabular_algs}
\reb{
We will detail the essential similarities and differences of the four algorithms \nv, Nash Q-Learning, \golf and \nd from four aspects: model-based/model-free, update manner, replay buffer, and exploration method.
\begin{itemize}
    \item \nv: model-based; update using full batch, no soft update; there is a buffer containing all samples so far;  $\epsilon$-greedy exploration.
    \item Nash Q-Learning: model-free; update using stochastic gradient for each sample, using soft update $Q\leftarrow (1-\alpha) Q + \alpha Q_\text{target}$;no replay buffer; $\epsilon$-greedy exploration.
    \item \golf: model-based; using an optimistic way of updating policy and exploiter within a confidence set; there is a buffer containing all samples so far;  a different behavior policy for exploration compared with $\epsilon$-greedy exploration.
    \item \nd: model-free; minibatch stochastic gradient update, using Mean Squared Error($Q, Q_\text{target}$) for gradient-based update; there is a buffer containing all samples so far;  $\epsilon$-greedy exploration.
\end{itemize}
From these similarities and differences, we can see that three theoretical algorithms Nash-VI, Nash Q-learning and GOLF-with-exploiter have slight differences in details, Nash-DQN can be viewed as practical approximation of both Nash-VI and Nash Q-learning.}

\section{Hyperparameters for Tabular Markov Game.}
\label{app:hyperparam_tabular}
\begin{table}[htbp]
\centering
\caption{Hyperparameters in Tabular Markov Game.}
\label{tab:rl_params}
\begin{tabular}{ccc}
\toprule
                           & \multirow{1}{*}{Hyperparameter} & \multicolumn{1}{c}{Values} \\ \cline{2-3} 
                          
\multirow{12}{*}{Common}       & Learning rate                   & $1\times 10^{-4}$ \\ \cline{2-3} 
                            & Optimizer                  & Adam            \\ \cline{2-3} 
                           & Batch size                      & $640$            \\ \cline{2-3} 
                           & Replay Buffer Size              & $10^5$            \\ \cline{2-3} 
                           & Episodes                        & $50000$              \\ \cline{2-3}
                           & Episode Length                  & $3$ for I / $6$ for II               \\ \cline{2-3} 
                           & Hidden Dimension                   & $128$             \\ \cline{2-3} 
                           & Hidden Activation                   & ReLU             \\ \cline{2-3} 
                           & Hidden Layers                   & $3$             \\ \cline{2-3} 
                           & Target Update Interval                   & 1000             \\ \cline{2-3} 
                           & $\gamma$               & $1.0$            \\ \cline{2-3} 
                           & $\epsilon$ Exponential Decay            & $\epsilon_0=1.0, \epsilon_1=0.0, p=8000$ \\ \midrule
\multirow{1}{*}{SP} &  $\delta$                 & $1.5$ for I / $2.0$ for II \\ \midrule
\multirow{1}{*}{FSP} &  $\delta$                 & $1.5$ for I / $2.0$ for II \\ \midrule
\multirow{1}{*}{NFSP} &  $\eta$                 & $0.1$ \\ \midrule 
\multirow{1}{*}{PSRO} &  $\delta$                 & $1.5$ for I / $2.0$ for II \\ \midrule
\multirow{1}{*}{Nash DQN Exploiter} 
 & Exploiter Update Ratio $m_2/m_1$ & 1 \\ \midrule
\end{tabular}
\end{table}

This section provides detailed hyperparameters of methods \textbf{with} function approximation on the tabular Markov games, as shown in Table~\ref{tab:rl_params}.
Methods such as SP, FSP, NFSP, PSRO all use DQN as the basic RL agent, and the ``Common'' hyperparameters are applied on the DQN algorithm. For \nd and \nde, since the algorithms follows a similar routine as DQN in general (value-based, off-policy), they also adopt the same hyperparameters.

In the common hyperparameters, the basic agent applies a network with $3$ hidden layers and $128$ as hidden dimension. The target update interval is the delayed update of the target network, and it updates once per $n$ times of standard network updates. $n$ is specified by the target update interval value, and $\gamma$ is the reward discounting coefficient, and it's set to be one in tabular Markov games since the episode length (3 for I or 6 for II) is small in the experiments. $\epsilon$ is the factor in $\epsilon$-greedy exploration and it follows an exponential decay schedule. Specifically, its value follows $\epsilon(t) = \epsilon_1 + (\epsilon_0 - \epsilon_1)e^{-t/p} $ in our experiments, where $t$ is the timestep in update. 

Since SP, FSP, PSRO algorithms follow an iterative best response procedure in the learning process, the margin for determining whether the current updating side achieves an approximate best response is set according to the average episodic reward. Once a learning agent wins over its opponent by a average reward threshold $\delta$ (depending on games), it saves the approximate best-response strategy and transfers the role to its opponent. The values of $\delta$ are different for the two tabular environments I and II, since the ranges of the return are different for the two environments. The longer horizon indicates a potentially larger range of return. 

For NFSP, since actions can be sampled from either $\epsilon$-greedy policy or the average policy, $\eta$ is a hyperparameter representing the ratio of choosing actions from $\epsilon$-greedy policy.

For \nde, the exploiter update ratio is the times of GD for the exploiter over the times GD for updating the Nash $Q$ network, which is $m_2/m_1$ as in Algorithm~\ref{alg:nash_dqn_exp}.

The exploiter for exploitation test after model training is a DQN agent with exactly the same ``Common'' hyperparameters for the tabular Markov game test.

\section{Results for Tabular Markov Games}
\label{sec:tabular_more_results}
Table~\ref{tab:tabular_exploit_compare128_var} shows the exploiter rewards in the exploitation test on two tabular Markov games. The exploiter reward is an approximation of $-V^{\hat{\mu},\dagger}(s_1)$ by Eq.~\eqref{eq:subopt}. The last column ``Nash V'' is the true value of $V^{\mu^*,\nu^*}(s_1)$. $V^{\mu^*,\nu^*}(s_1)-V^{\hat{\mu},\dagger}(s_1)$ gives the true exploitability for $\hat{\mu}$.
Due to the randomly generated payoff matrix, it is asymmetric for the two players: $V^{\mu^*,\nu^*}(s_1)\approx-0.296$ for environment I and $V^{\mu^*,\nu^*}(s_1)\approx-0.131$ for environment II.   For example, in Table~\ref{tab:tabular_exploit_compare128_var} the mean of approximate exploitability of Oracle Nash is $0.269-0.296=-0.027$, and the theoretical value should be zero since it is the ground truth Nash equilibrium strategy. Also, since the transition and reward is assumed to be unknown and the exploitability is approximated with a DQN agent, the stochasticity of the results is larger than the tabular method test. Table~\ref{tab:tabular_exploit_compare128_multi_stages} shows the exploitation results at different training stages for environment I. At step 0, some methods have common results since the initialized models are the same for them.

\begin{table*}[htbp]
\centering
\caption{Exploiter rewards in tabular case: the reward means and standard deviations are derived over 1000 episodes in exploitation test. }
\resizebox{\textwidth}{!}{ 
\begin{tabular}{ c|c|c|c|c|c|c|c||c } 
\hline
Env / Method & SP & FSP & NFSP & PSRO & Nash DQN & Nash DQN Exploiter & Oracle Nash & Nash V\\
\hline
Tabular Env I & $0.744\pm{0.711}$ & $0.675\pm{0.811}$ & $0.613\pm{0.858}$ & $0.430\pm{1.110}$ & $0.392\pm{1.044}$ & $0.316\pm{0.998}$ & $0.269\pm{1.029}$ & $-0.296$ \\ \hline
Tabular Env II & $1.370\pm1.323$ & $0.825\pm 1.343$ & $0.510\pm1.408$ & $0.700\pm1.337$ & $0.148\pm1.341$ & $ 0.202\pm1.425$ & $0.049\pm1.365$ & $-0.131$\\
\hline
\end{tabular}
}
\label{tab:tabular_exploit_compare128_var}
\end{table*}

\begin{table*}[htbp]
\centering
\caption{Exploiter rewards in tabular Markov game I after training for 0, 10k, 20k, 30k, 40k, 50k episodes and exploitation for 30k episodes. The reward means and standard deviations are derived over 1000 episodes in exploitation test.}
\resizebox{\textwidth}{!}{ 
\begin{tabular}{c|c|c|c|c|c|c}
\toprule
\multirow{2}{*}{Method} & \multicolumn{6}{c}{Exploiter Reward} \\
\cline{2-7}
 & \multicolumn{1}{c|}{0}  & \multicolumn{1}{c|}{10k} & \multicolumn{1}{c|}{20k} & \multicolumn{1}{c|}{30k} &  \multicolumn{1}{c|}{40k} & \multicolumn{1}{c}{50k}\\
\hline
 SP & \multirow{4}{*}{$1.114\pm{0.776}$} & $1.133\pm{0.764}$ & $1.002\pm{0.786}$ & $0.749\pm{0.907}$ & $0.885\pm{0.837}$ & $0.744\pm{0.711}$\\ \cline{1-1} \cline{3-7}
 FSP & & $0.724\pm{0.890}$ & $0.698\pm{0.808}$ & $0.690\pm{0.783}$ & $0.709\pm{0.801}$ & $0.675\pm{0.811}$\\ \cline{1-1} \cline{3-7}
 NFSP & & $0.607\pm{0.982}$ & $0.604\pm{0.887}$ & $0.610\pm{0.951}$ & $0.500\pm{0.933}$ & $0.613\pm{0.858}$ \\  \cline{1-1} \cline{3-7}
 PSRO & & $0.669\pm{0.873}$ & $0.650\pm{0.829}$ & $0.714\pm{0.790}$ & $0.712\pm{0.913}$ & $0.430\pm{1.110}$ \\ \hline
 Nash DQN & \multirow{2}{*}{$1.214\pm{0.860}$}  & $0.405\pm{1.003}$ & $0.392\pm{1.040}$ & $0.387\pm{1.003}$ & $0.411\pm{1.040}$ & $0.392\pm{1.044}$\\ \cline{1-1} \cline{3-7}
 Nash DQN Exploiter & & $0.301\pm{0.978}$ & $0.297\pm{1.056}$ & $0.273\pm{1.034}$ & $0.291\pm{0.945}$ & $0.316\pm{0.998}$ \\ \hline
 Oracle Nash & \multicolumn{6}{c}{$0.269\pm{1.029}$}\\
\bottomrule
\end{tabular}
}
\vskip -.1in
\label{tab:tabular_exploit_compare128_multi_stages}
\end{table*}


\section{Hyperparameters for Two-Player Video Games.}
\label{app:hyperparam_nontabular}

\begin{table}[htbp]
\centering
\caption{Hyperparameters in SlimeVolley.}
\label{tab:rl_params_sv}
\begin{tabular}{ccc}
\toprule
                           & \multirow{1}{*}{Hyperparameter} & \multicolumn{1}{c}{Values} \\ \cline{2-3} 
                          
\multirow{12}{*}{Common}       & Learning rate                   & $1\times 10^{-4}$ \\ \cline{2-3} 
                            & Optimizer                  & Adam            \\ \cline{2-3} 
                           & Batch size                      & $128$            \\ \cline{2-3} 
                           & Replay Buffer Size              & $10^5$            \\ \cline{2-3} 
                           & Episodes                        & $50000$              \\ \cline{2-3}
                           & Episode Length                  & $\le300$              \\ \cline{2-3} 
                           & Hidden Dimension                   & $128$             \\ \cline{2-3} 
                           & Hidden Layers                   & $3$             \\ \cline{2-3} 
                           & Hidden Activation                   & ReLU             \\ \cline{2-3} 
                           & Target Update Interval                   & 1000             \\ \cline{2-3} 
                           & $\gamma$               & $0.99$           \\ \cline{2-3} 
                           & $\epsilon$ Exponential Decay            & $\epsilon_0=1.0, \epsilon_1=1\times10^{-3}, p=1\times 10^5$  \\ \midrule
\multirow{1}{*}{SP} &  $\delta$                 & $3.0$ \\ \midrule
\multirow{1}{*}{FSP} &  $\delta$                 & $3.0$ \\ \midrule
\multirow{1}{*}{NFSP} &  $\eta$                 & $0.1$ \\ \midrule 
\multirow{1}{*}{PSRO} &  $\delta$                 & $3.0$ \\ \midrule
\multirow{1}{*}{Nash DQN} &  $\epsilon$ Exponential Decay            & $\epsilon_0=1.0, \epsilon_1=1\times10^{-3}, p=5\times 10^6$  \\ \midrule
\multirow{2}{*}{Nash DQN Exploiter} &  $\epsilon$ Exponential Decay            & $\epsilon_0=1.0, \epsilon_1=1\times10^{-3}, p=5\times 10^6$  \\ \cline{2-3}
 & Exploiter Update Ratio $m_2/m_1$ & 1 \\ \midrule
                         
\end{tabular}
\end{table}

\begin{table}[htbp]
\centering
\caption{Hyperparameters in two-player Atari games.}
\label{tab:rl_params_atari}
\begin{tabular}{ccc}
\toprule
                           & \multirow{1}{*}{Hyperparameter} & \multicolumn{1}{c}{Values} \\ \cline{2-3} 
                          
\multirow{12}{*}{Common}       & Learning rate                   & $1\times 10^{-4}$ \\ \cline{2-3} 
                            & Optimizer                  & Adam            \\ \cline{2-3} 
                           & Batch size                      & $128$            \\ \cline{2-3} 
                           & Replay Buffer Size              & $10^5$            \\ \cline{2-3} 
                           & Episodes                        & $50000$              \\ \cline{2-3}
                           & Episode Length                  & $\le300$      \\ \cline{2-3} 
                           & Hidden Dimension                   & $128$             \\ \cline{2-3} 
                           & Hidden Layers                   & $4$             \\ \cline{2-3} 
                           & Hidden Activation                   & ReLU             \\ \cline{2-3} 
                           & Target Update Interval                   & 1000             \\ \cline{2-3} 
                           & $\gamma$               & $0.99$            \\ \cline{2-3} 
                           & $\epsilon$ Exponential Decay            & $\epsilon_0=1.0, \epsilon_1=1\times10^{-3}, p=1\times 10^5$  \\ \midrule
\multirow{1}{*}{SP} &  $\delta$                 & $80/15/10/7/3$ \\ \midrule
\multirow{1}{*}{FSP} &  $\delta$                 & $80/15/10/7/3$ \\ \midrule
\multirow{1}{*}{NFSP} &  $\eta$                 & $0.1$ \\ \midrule 
\multirow{1}{*}{PSRO} &  $\delta$                 & $80/15/10/7/3$ \\ \midrule
\multirow{1}{*}{Nash DQN} &  $\epsilon$ Exponential Decay            & $\epsilon_0=1.0, \epsilon_1=1\times10^{-3}, p=5\times 10^6$  \\ \midrule
\multirow{2}{*}{Nash DQN Exploiter} &  $\epsilon$ Exponential Decay            & $\epsilon_0=1.0, \epsilon_1=1\times10^{-3}, p=5\times 10^6$  \\ \cline{2-3}
 & Exploiter Update Ratio $m_2/m_1$ & 3 \\ \midrule
                         
\end{tabular}
\end{table}
This section provides detailed hyperparameters of methods \textbf{with} function approximation on the tabular Markov games, as shown in Table~\ref{tab:rl_params_sv} for \textit{SlimeVolley} environment and Table~\ref{tab:rl_params_atari} for two-player Atari games.
For Table~\ref{tab:rl_params_sv} and \ref{tab:rl_params_atari}, the meaning of each hyperparameter is the same as in Appendix Sec.~\ref{app:hyperparam_tabular}.
For Table~\ref{tab:rl_params_atari}, the threshold value $\delta$ for iterative best response procedure has multiple values, which correspond in order with environments \textit{Boxing-v1}, \textit{Double Dunk-v2}, \textit{Pong-v2}, \textit{Tennis-v2}, \textit{Surround-v1}, respectively. 

It can be noticed that \nd and \nde use a different $\epsilon$ decay schedule from other methods including SP, FSP, NFSP and PSRO. Since the Nash-based methods follow a single model update routine, the slow decaying $\epsilon$ (larger $p$) is more proper to be applied. Other baseline methods follow the iterative best response procedure, which learns new models in each period of update iteratively. Each period within the overall training process is much shorter, therefore the model with a faster decaying $\epsilon$ will learn better. In between two periods, the $\epsilon$ is re-initialized as the starting value $\epsilon_0$ during the whole training process.

The exploiter for exploitation test after model training is a DQN agent with exactly the same common hyperparameters for the two-player video game test.

\section{Complete Results for Video Games}
\label{app:sec_results_video}

Table~\ref{tab:atari_exploit_best} shows the means of exploitation results over different runs and exploiters. The experiments are the same as in Table~\ref{tab:atari_exploit}. Notice that the mean values of Table~\ref{tab:atari_exploit_best} are different from the values in Table~\ref{tab:atari_exploit}. This is because in Table~\ref{tab:atari_exploit_best} the models with best unexploitable performance are reported, while in Table~\ref{tab:atari_exploit} it's averaged over all runs and exploitation tests. 

\begin{table*}[htbp]
\scriptsize
\centering
\caption{Approximate exploitability (lower is better) for six two-player video games.}
\resizebox{\textwidth}{!}{ 
\begin{tabular}{>{\centering\arraybackslash}m{60pt}|>{\centering\arraybackslash}m{30pt}|>{\centering\arraybackslash}m{30pt}|>{\centering\arraybackslash}m{30pt}|>{\centering\arraybackslash}m{30pt}|>{\centering\arraybackslash}m{47pt}|>{\centering\arraybackslash}m{89pt}}
\toprule
\multicolumn{1}{c|}{\backslashbox{Env}{Method}} & 
 \multicolumn{1}{c|}{SP}  & \multicolumn{1}{c|}{FSP} & \multicolumn{1}{c|}{NFSP} & \multicolumn{1}{c|}{PSRO} &  \multicolumn{1}{c|}{\textbf{Nash DQN}} & \multicolumn{1}{c}{\textbf{Nash DQN Exploiter}}\\
\hline
  SlimeVolley & $-0.049$ & $0.514$ & $0.069$ & $0.000$ & $\cellcolor{gray!15}0.000$ & $\cellcolor{gray!15}\color{blue}\mathbf{-0.099}$\\ \hline
 Boxing & $24.907$ & $93.683$ & $24.544$ & $66.891$ & $\cellcolor{gray!15}\color{blue}\mathbf{-55.471}$ & $22.490$\\  \hline
 Double Dunk & $7.039$ & $6.067$ & $4.564$ & $7.256$ & $\cellcolor{gray!15}\color{blue}\mathbf{-0.539}$ & $1.702$ \\  \hline
 Pong & $4.207$ & $5.196$ & $4.396$ & $5.217$ & $\cellcolor{gray!15}\color{blue}\mathbf{-3.336}$ & $\cellcolor{gray!15}-2.920$ \\ \hline
 Tennis  & $2.970$ & $2.355$ & $3.207$ & $2.465$ & $\cellcolor{gray!15}\color{blue}\mathbf{-0.425}$ & $0.069$ \\  \hline
 Surround  & $1.782$ & $1.574$ & $1.594$ & $1.603$ & $\cellcolor{gray!15}\color{blue}\mathbf{0.904}$ & $1.462$ \\ 
\bottomrule
\end{tabular}
}
\vskip -.1in
\label{tab:atari_exploit_best}
\end{table*}



\begin{figure}[htp]
    \centering
    \includegraphics[height=0.95\textheight, width=\columnwidth]{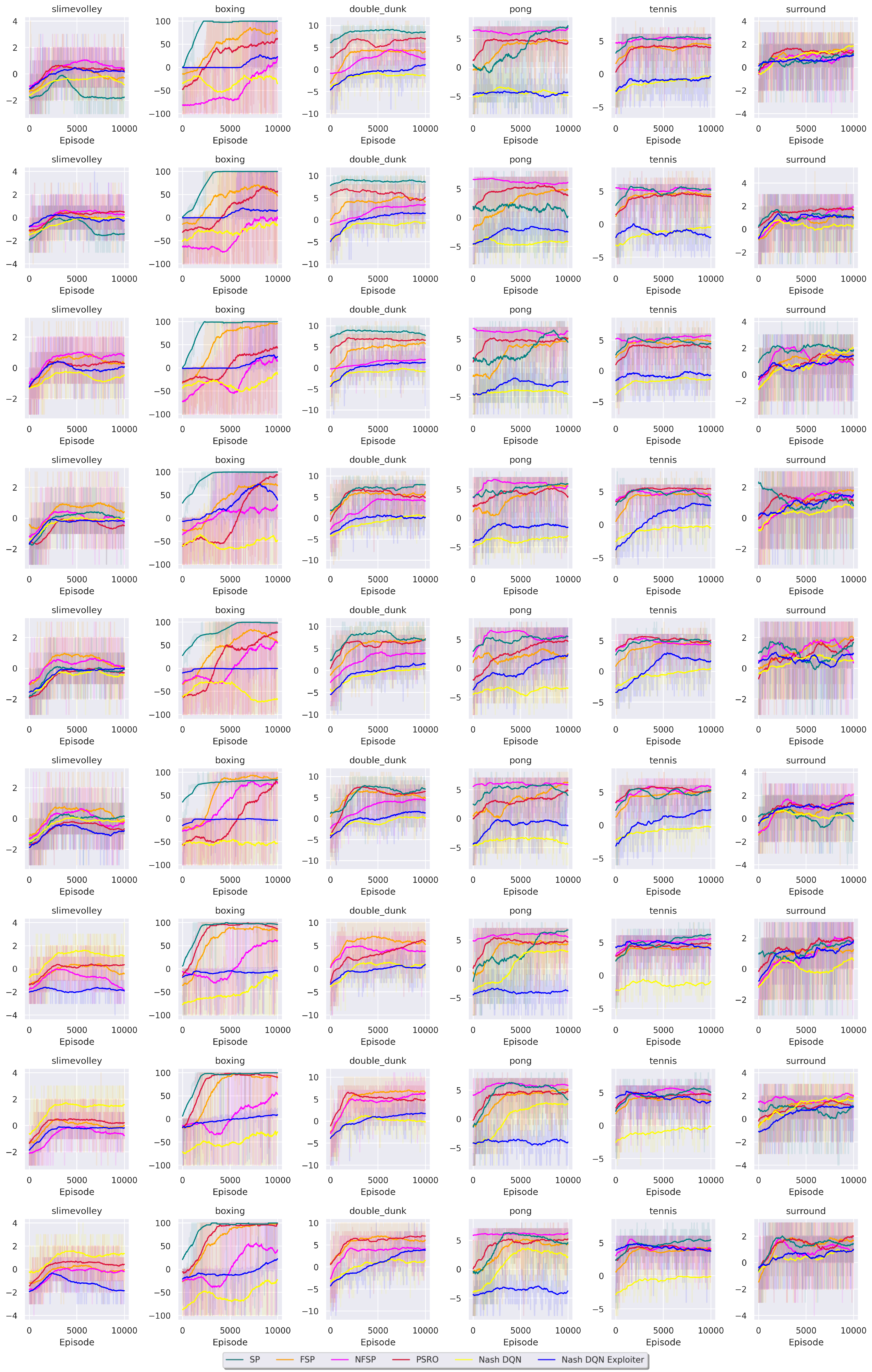}
    \caption{The exploitation tests on the first-player side on all six two-player zero-sum video games, for three runs and three exploitation for each run. The vertical axis is the episodic exploiter reward. }
    \label{fig:exploit_all_first}
\end{figure}

\begin{figure}[htp]
    \centering
    \includegraphics[height=0.95\textheight, width=\columnwidth]{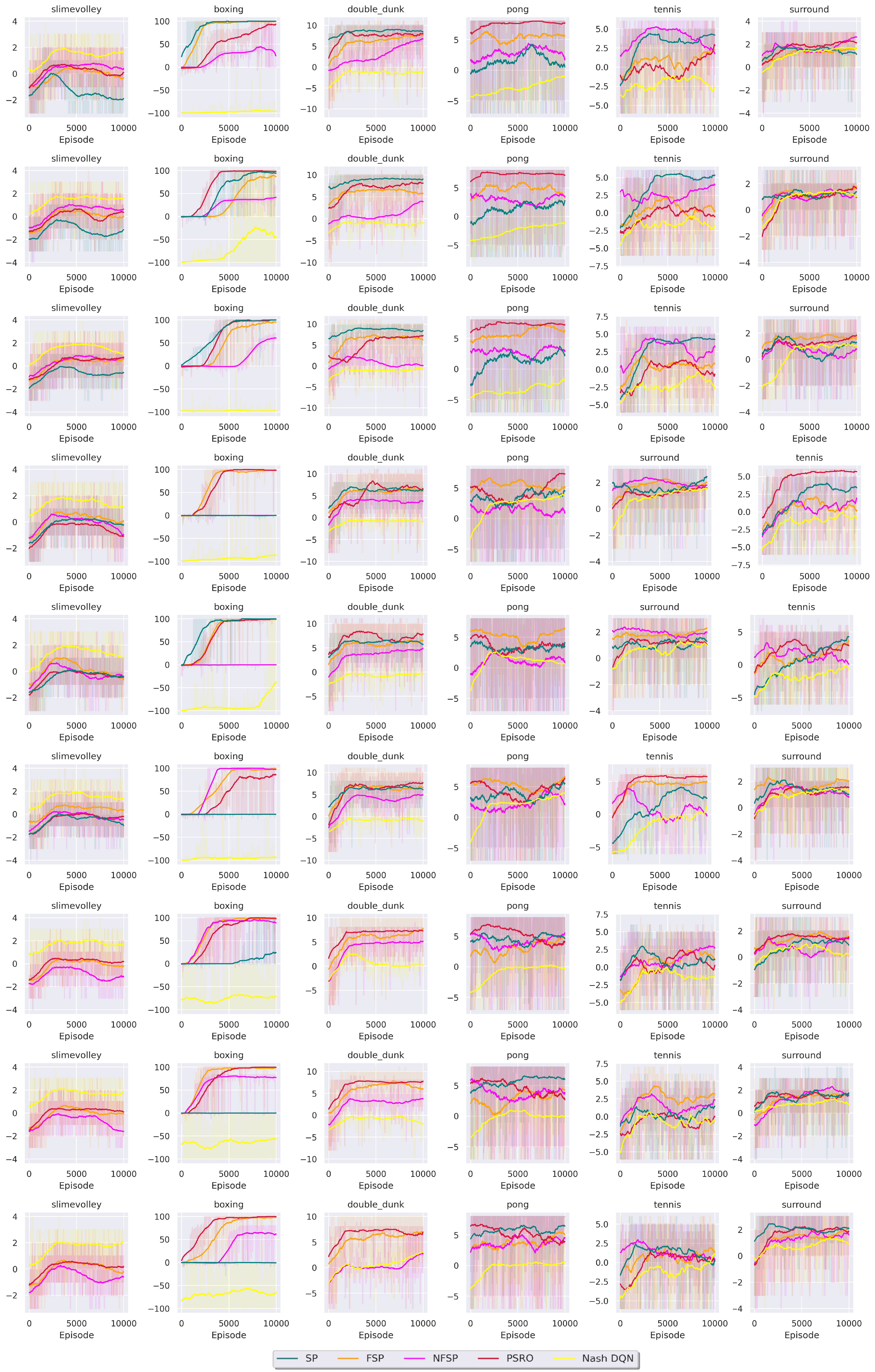}
    \caption{The exploitation tests on the first-player side on all six two-player zero-sum video games, for three runs and three exploitation for each run. The vertical axis is the episodic exploiter reward. }
    \label{fig:exploit_all_second}
\end{figure}

All results for three runs and each with three exploitation tests are shown in Fig.~\ref{fig:exploit_all_first} and \ref{fig:exploit_all_second}, which corresponding to the exploitation of the first and second player in games respectively. The \nde method is asymmetric and the second player side is not the NE strategy, so in Fig.~\ref{fig:exploit_all_second} there is no exploitation results for \nde. All experiments are conducted on a 8-GPU (Nvidia Quadro RTX A6000 48GB) server with 192 CPU cores. The exploitation test is evaluated with non-greedy DQN agent for one episode every 20 training episodes during the whole training period. All curves are smoothed with a window size of $100$.

\reb{Notice that in above experiments the exploitability value can be negative sometimes, however Eq.~\ref{eq:subopt} tells the exploitability for symmetric games should always be non-negative. This is because Eq.~\ref{eq:subopt} shows the theoretical best response, which is not tractable for large-scale games like Atari. The data we reported tables are exploitability results approximated with single-agent RL (DQN). We expect these exploiters to be weaker than the theoretically optimal ones, therefore negative values could appear, which also indicates that the learned Nash-DQN agents are extremely difficult to be exploited.}

\section{Experiments for Full Length Environments}
\label{app:sec_full_len}
\begin{figure}[H]
    \centering
    \includegraphics[width=0.6\columnwidth]{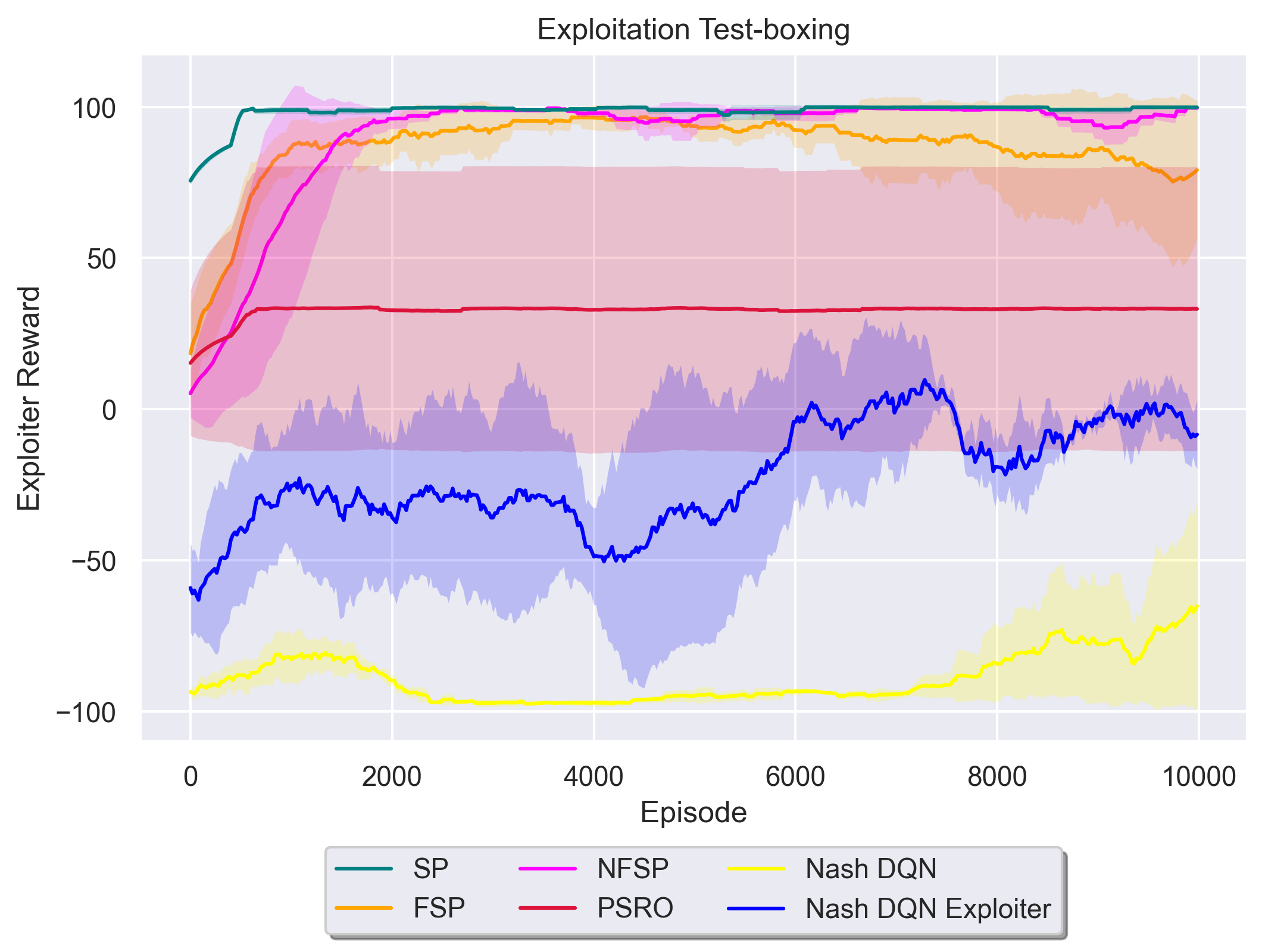}
    \caption{Comparison of the exploiter learning curves for a full-length  setting on \textit{Boxing-v1}.}
    \label{fig:full_len_env}
\end{figure}

In Fig.\ref{fig:full_len_env}, we show the exploitability experimentation with the full-length environment \textit{Boxing-v1} (no truncation to 300 steps), and also the experiments with the same settings as  in Sec.~\ref{sec:video_game} except for the episode length. As shown in the figure, our methods (yellow for \nd and blue for \nde) show the best exploitability performance, which are less exploited in the exploitation tests. Specifically, even in this full length experimentation, the best performance method \nd achieves non-exploitable strategies.

\end{document}